%% file: AutoFed.tex
\documentclass[sigconf]{acmart}

\usepackage[noend]{algorithm2e}
\usepackage{subfig}
\usepackage{amsmath}
\usepackage{lipsum}

\DeclareMathOperator*{\argmin}{arg\,min}

\ifodd 0
\newcommand{\brev}[1]{{\color{blue}#1}}

\else
\newcommand{\brev}[1]{#1}

\fi

\ifodd 0
\newcommand{\newrev}[1]{{\color{red}#1}}    %
\newcommand{\needrev}[1]{{\color{green}#1}} %
\else
 
\newcommand{\newrev}[1]{#1} 
\newcommand{\needrev}[1]{#1} 
\fi 

\AtBeginDocument{%
  \providecommand\BibTeX{{%
    Bib\TeX}}}

\setcopyright{acmcopyright}
\copyrightyear{2023}
\acmYear{2023}
\acmDOI{XXX.XXX}
\acmConference[ACM MobiCom'23]{The 29th Annual International Conference on Mobile Computing and Networking}{October 2-6, 2023}{Madrid, Spain}
\acmBooktitle{The 29th Annual International Conference on Mobile Computing and Networking (ACM MobiCom'23), October 2-6, 2023, Madrid, Spain}
\acmPrice{15.00}
\acmISBN{978-1-4503-XXXX-X/18/06}

\AtBeginDocument{%
  \providecommand\BibTeX{{%
    \normalfont B\kern-0.5em{\scshape i\kern-0.25em b}\kern-0.8em\TeX}}}

\begin{document}

\newcommand{\sysname}{AutoFed}

\title{AutoFed: Heterogeneity-Aware Federated Multimodal Learning for Robust Autonomous Driving}

\ifodd 0
\author{
	Conditional Accepted Paper \#199 to ACM MobiCom 2023
    }
    \thanks{*~***. \vspace{-1ex}}
\affiliation{
    \institution{\color{white}
	$^1$ABA \country{AA} \\ 
	$^2$BAB \country{BB} \\
	$^3$BAB \country{CC} \\
    Email: aaa@bbb.cc
        }
    }
    \renewcommand{\authors}{Conditionally Accepted Paper \#199 to ACM MobiCom 2023}
    \renewcommand{\shortauthors}{Conditionally Accepted Paper \#199 to ACM MobiCom 2023}
\else
\author{ 
    Tianyue Zheng$^{1}$\quad Ang Li$^2$\quad Zhe Chen$^{3*}$\quad Hongbo Wang$^{1}$\quad Jun Luo$^{1}$
    }
    \thanks{* is the correspondence author. © {Tianyue Zheng, Ang Li, Zhe Chen, Hongbo Wang, and Jun Luo | ACM} {2023}. This is the author's version of the work. It is posted here for your personal use. Not for redistribution. The definitive Version of Record is accepted by ACM MobiCom 2023. }
\affiliation{
   \institution{
	$^1$School of Computer Science and Engineering, Nanyang Technological University (NTU) \country{Singapore} \\
	$^2$Department of Electrical and Computer Engineering, University of Maryland \country{USA} \\ 
	$^3$AIWiSe, China-Singapore International Joint Research Institute \country{China} \\
    Email: \{tianyue002, junluo\}@ntu.edu.sg, angli@umd.edu,
    chenz@csijri.com
        }
    }
    \renewcommand{\authors}{T. Zheng, A. Li, Z. Chen, H, Wang, and J. Luo}
    \renewcommand{\shortauthors}{T. Zheng, A. Li, Z. Chen, H, Wang, and J. Luo}
\fi

\renewcommand{\shorttitle}{\sysname}

\begin{abstract}
 Object detection with on-board sensors (e.g., lidar, radar, and camera) play a crucial role in \textit{autonomous driving} (AD), and these sensors complement each other in modalities. While crowdsensing may potentially exploit these sensors (of huge quantity) to derive more comprehensive knowledge, \textit{federated learning} (FL) appears to be the necessary tool to reach this potential: it enables \textit{autonomous vehicle}s (AVs) to train machine learning models without explicitly sharing raw sensory data. However, the multimodal sensors introduce various data heterogeneity across distributed AVs (e.g., label quantity skews and varied modalities), posing critical challenges to effective FL. To this end, we present \textbf{AutoFed} as a heterogeneity-aware FL framework to fully exploit multimodal sensory data on AVs and thus enable robust AD. Specifically, we first propose a novel model leveraging pseudo labeling to avoid mistakenly treating unlabeled objects as the background. We also propose an autoencoder-based data imputation method to fill missing data modality (of certain AVs) with the available ones. To further reconcile the heterogeneity, we finally present a client selection mechanism exploiting the similarities among client models to improve both training stability and convergence rate. Our experiments on benchmark dataset confirm that AutoFed substantially improves over status quo approaches in both precision and recall, while demonstrating strong robustness to adverse weather conditions.
\end{abstract}



\maketitle

\input{1_introduction}
\input{2_background_motivation}

\input{3_design}

\input{4_evaluation}
\input{6_related_work}

\vspace{-.5ex}
\input{7_conclusion}

\balance
\bibliographystyle{ACM-Reference-Format}

\end{document}

%% file: 1_introduction.tex
\section{Introduction}\label{sec:introduction}

Undergoing worldwide rapid development~\cite{tesla, waymo, uber}, \textit{autonomous driving} (AD) aims to offer a wide range of benefits including better safety, less harmful emissions, increased lane capacity, and less travel time~\cite{shaheen2019mobility}. The core of AD is the perception capability to detect objects (e.g, vehicles, bicycles, signs, pedestrians) on the road; it enables interpretable path and action planning. Formally, the SAE (Society of Automotive Engineers) requires Level 3-5 AD to be able to monitor environments and detect 
objects, even under adverse road and weather conditions~\cite{sae2014taxonomy}. 
\brev{To reach these goals, \textit{multiple on-board sensing modalities} (e.g., lidar, radar, and camera) collaboratively deliver complementary and real-time information of the surroundings. While lidar and camera provide high-definition measurements in short distance due to
attenuation in distance and degradation by adverse weather or lighting conditions,
radar achieves relatively longer-range monitoring robust to adverse conditions, leveraging the penetrating power of radio waves~\cite{zheng2021siwa}.} 

\brev{To fully take advantage of the rich multimodal information provided by various sensors, a plethora of previous arts~\cite{ku2018joint, xu2018pointfusion, qi2018frustum, chen2017multi, liang2018deep, qian2021robust,liu2021bev} have employed deep learning to perform multi-modality fusion as well as pattern recognition, aiming to conduct accurate and reliable \textit{object detection} (OD). 
The mainstream of OD relies on a two-stage method~\cite{ku2018joint, xu2018pointfusion, qi2018frustum, chen2017multi, qian2021robust}, where proposals for regions of interest are generated first and then refined for object classification and bounding box regression. 
Though OD can handle different viewing angles in general, we focus only on \textit{bird's-eye view}~\cite{qian2021robust, liu2021bev} for reduced complexity, as it 
reconciles the view discrepancy among different sensing modalities at a reasonably low cost.
Yet even with this cost reduction, the fundamental difference between OD and basic learning tasks (e.g., classification) still lead to far more (deep learning) model parameters than normal, rendering its training hard to converge even for a single model, yet we shall further promote the need for training multiple models in a distributed manner.}

Ideally, deep neural networks (DNNs) for OD should be trained on a dataset that takes into account different road, traffic, and weather conditions. However, the ever-changing driving environments render it infeasible for car manufacturers and developers to collect a dataset covering all scenarios. Whereas crowdsensing~\cite{ganti2011mobile, incentive, Dual-DCOSS14, GROPING-TMC15, Quality-ToN18, Learning-TMC19} can be exploited to overcome this difficulty by outsourcing data collection and annotation tasks to \textit{autonomous vehicle}s (AVs), conventional crowdsensing suffers from privacy concerns~\cite{vergara2016privacy,de2013unique} and data communication burdens. Fortunately, 
integrating \textit{federated learning} (FL)~\cite{konevcny2016federated} into crowdsensing could virtually tackle these problems. As an emerging paradigm for distributed training across massive participating clients,
FL demands a central \textit{server} only to coordinate the distributed learning process, with which each \textit{client} shares only the local model parameters: such a scheme protects data privacy while reducing communication load at the same time. 

\textbf{Challenges.} Designing an FL system for OD upon AVs' multimodal data is challenging, because all three entities (i.e., human, vehicle, and environment) involved in the system introduce a variety of data heterogeneity. First, relying on clients to label the data can lead to annotation heterogeneity: some clients may be more motivated to provide 
annotations with adequate quality (e.g., bounding boxes around the detected vehicle), while others may be so busy and/or less skillful that they miss a large proportion of the annotations. Second, crowdsensing by different AVs also introduces sensing modality heterogeneity,\footnote{\brev{Training a single model using FL under the heterogeneity of sensing modality allows vehicles with fewer sensors to learn from others via FL, and it also improves the model robustness against sensor malfunctions.}} since the vehicles may be equipped with different types of sensors by their manufacturers. Even for AV models from the same manufacturer, it is common that certain sensor experiences malfunction, causing corresponding data modality to get lost.
Third, the ever-changing environment (e.g, weather and road) can introduce drifts in data distributions, further exacerbating the heterogeneity issue. Unfortunately, prior arts on FL either focus on homogeneous scenarios~\cite{mcmahan2017communication, konevcny2016federated}, or deal with heterogeneity of unimodal data~\cite{li2020federated, fallah2020personalized, tu2021feddl}). None of existing works is capable of handling all the aforementioned heterogeneity induced by humans, vehicle, and environment under our targeted AD scenarios. \brev{Last but not least, the high complexity of the two-stage OD network makes its loss surface chaotic~\cite{li2018visualizing}, further exacerbating the negative impacts of the data heterogeneity on the performance.}

\begin{figure}[t]
    \setlength\abovecaptionskip{8pt}
    \centering
	\includegraphics[width=\linewidth]{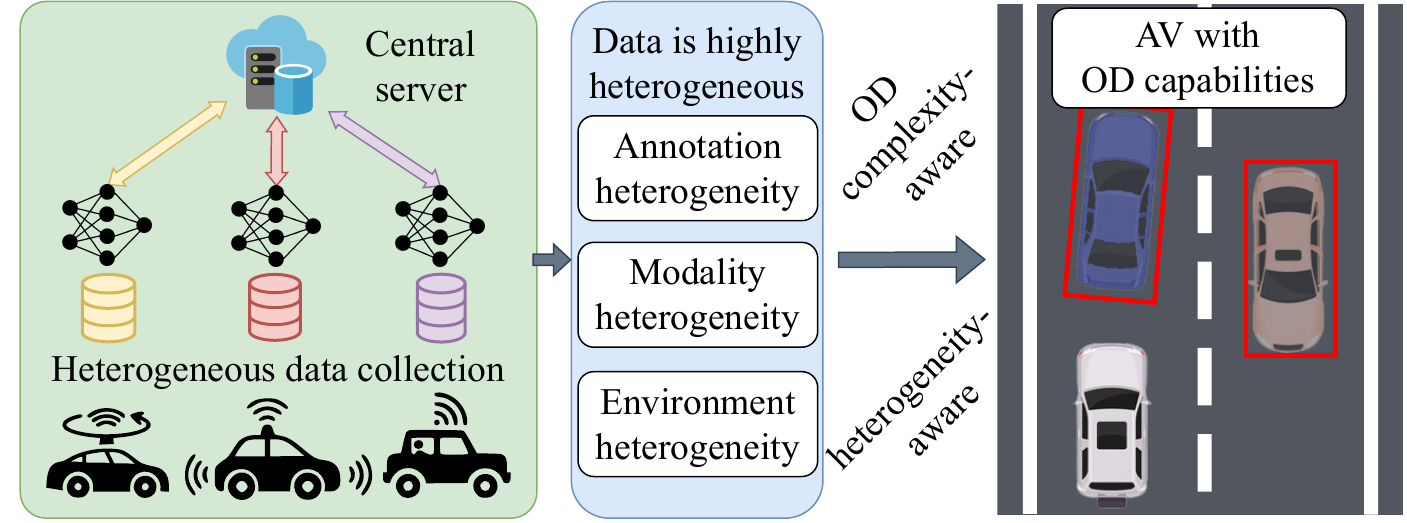}
	\caption{\brev{The bird's-eye view FL-OD of \sysname.}}
	\label{fig:teaser}
\end{figure}

\textbf{Our solutions.} To tackle these challenges, \brev{we carefully re-engineer the classical two-stage OD model~\cite{ren2015faster} to accommodate the multimodal data generated on AVs, and we exploit a major insight on the loss surface of OD under FL to guide the design of several learning mechanisms that handle the heterogeneity issue, 
as briefly illustrated in Figure~\ref{fig:teaser}. 
Since we notice that tolerance for data anomalies is crucial to efficiently navigate on the chaotic loss surface, we focus on robust designs to achieve such tolerance. 
}
Specifically, we design a cross-entropy-based loss function for training the neural model to handle unlabeled regions (of certain vehicles) that could be mistakenly regarded as the background during training. \sysname\ also employs inter-modality autoencoders to perform data imputation of missing sensor modalities. The autoencoders learn from incomplete data modality and generate plausible values for the missing modality. 
Finally, \sysname\ exploits a novel client selection mechanism to \brev{handle environment heterogeneity by eliminating diverged models. All in all, these three mechanisms together may largely avoid data abnormality and hence prevent the clients' losses from falling into local minimums on the chaotic loss surface.} 
Our key contributions can be summarized as follows:

\begin{itemize}
    \item To the best of our knowledge, \sysname\ is the first FL system specifically designed for multi-modal OD under heterogeneous AV settings.
    \item We design a novel cross entropy loss for training the neural model for OD, aiming to mitigate the annotation heterogeneity across clients.
    \item We design an inter-modality autoencoder to perform missing  data modality imputation, thus alleviating the modality heterogeneity across the clients.
    \item We design a novel client selection mechanism for choosing mutually-enhancing clients, thus further eliminating the harmful effects induced by heterogeneity.
    \item We implement \sysname\ prototype and evaluate \sysname\ with extensive experiments. The promising results demonstrate that \sysname\ can enable robust vehicle detection under AD scenarios.
\end{itemize}
Whereas most FL proposals consider only basic learning tasks~\cite{konevcny2016federated, li2020federated, li2021fedmask, tu2021feddl}, \sysname\ pioneers in FL-driven AV-OD far more sophisticated yet realistic than basic classification or regression. 
In the following, \S~\ref{sec:background_motivation} motivates the design of \sysname\ by revealing the damaging effects of the heterogeneity. \S~\ref{sec:system_design} presents the system design of \sysname. \S~\ref{sec:evaluation} introduces the datasets, system implementation, and experiment setup, before reporting the evaluation results. Related works and technical limitations are discussed in \S~\ref{sec:related_work}. Finally, \S~\ref{sec: conclusion} concludes the paper with future directions.

%% file: 2_background_motivation.tex
\section{Motivation}\label{sec:background_motivation}

We first investigate the impact of annotation heterogeneity on the performance of a DNN model for vehicle detection. Then we show that the heterogeneous modality significantly degrades the performance of the federated model on OD. Finally, we confirm the necessity to tackle the model divergence potentially caused by heterogeneous factors (e.g., diversified environments and human inputs) in federated training.

\subsection{Quantity Skew of Labeled Data} \label{ssec:label_skew}
As an FL system, AutoFed relies on AV clients to provide labels (i.e., bounding boxes around vehicles) for two reasons: i) the server should not have access to local data due to privacy concerns; and ii) labeling data locally is more reasonable compared to performing the labeling offline on the server, \brev{because more visual cues can be leveraged for labeling locally on AVs,\footnote{\brev{The driver/co-pilot can provide online
labels similar to crowdsourcing in Waze~\cite{waze}.}} and we hence deem all such labels as reliable. However, relying on clients for data labeling can lead to skew in label quantity:} some clients may be more motivated to provide annotations with adequate quality, while others may be so busy and/or less skillful that they miss a large proportion of the annotations. The situation may get worse during training, as the missing annotations on some AVs could be mistakenly marked as background by the OD network, thus backpropagating wrong gradients during local training. As a result, the small number of labels on some AVs may degrade the overall performance and cause training instability of the OD network.  
\begin{figure}[t]
    \setlength\abovecaptionskip{6pt}
    \vspace{-2.5ex}
	   \captionsetup[subfigure]{justification=centering}
		\centering
		\subfloat[Average precision.]{
		  \begin{minipage}[b]{0.47\linewidth}
		        \centering
			    \includegraphics[width = 0.96\textwidth]{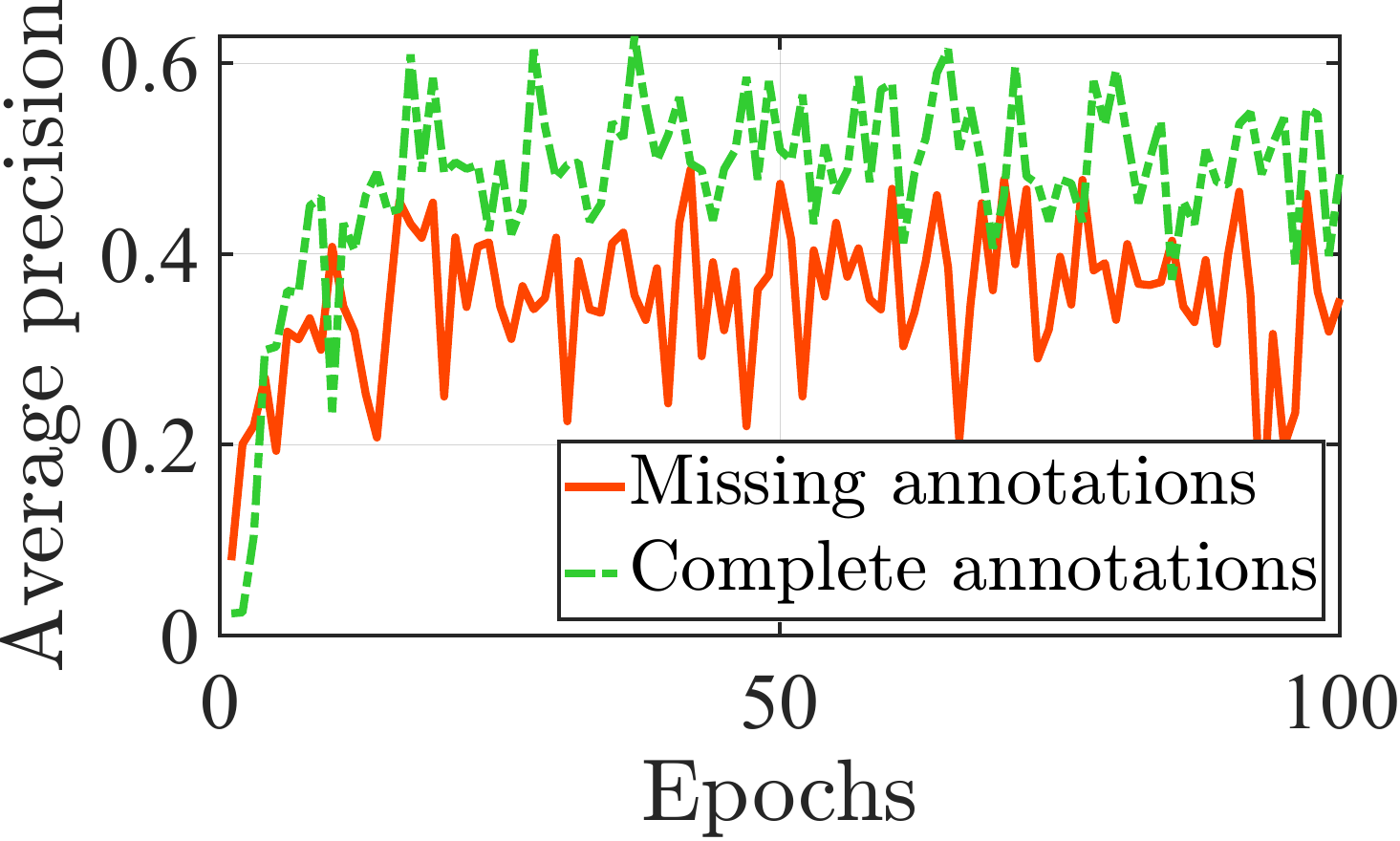}
			\end{minipage}
		}
		\subfloat[Average recall.]{
		    \begin{minipage}[b]{0.47\linewidth}
		        \centering
			    \includegraphics[width = 0.96\textwidth]{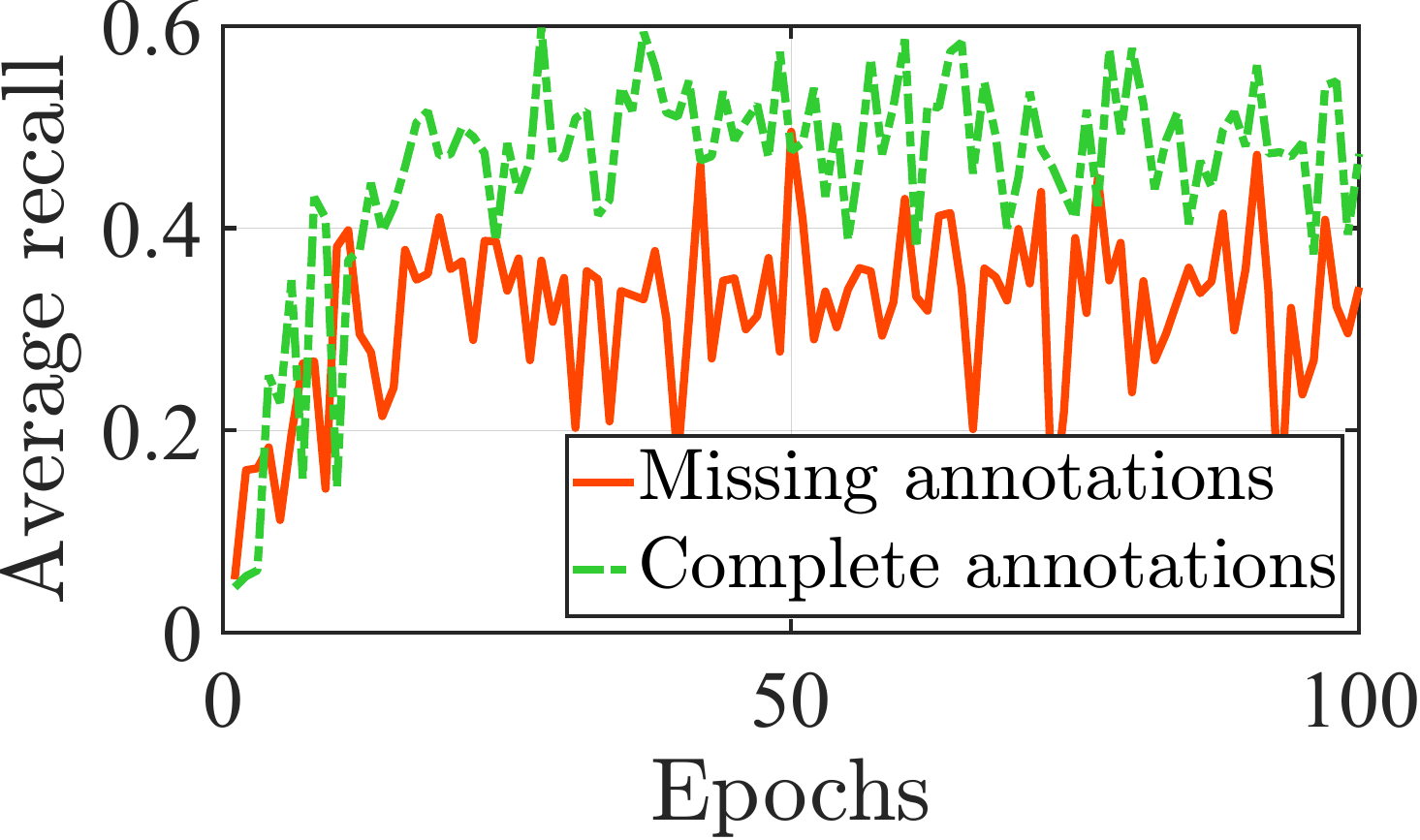}
			\end{minipage}
		}
		\caption{Damaging effects of missing annotations.}
		\label{fig:damage_miss_anno}
	    \vspace{-2ex}
\end{figure}
To demonstrate the damaging effects of missing labeling, we show the average precision and recall of %
\brev{a two-stage OD network for the task of vehicle detection in Figure~\ref{fig:damage_miss_anno}. The network utilizes a VGG variant~\cite{simonyan2014very} as its backbone and was trained by standard backpropagation using an SGD optimizer on a dataset of 1,000 data samples with 50\% data with missing annotations and 50\% data with complete annotations in a standalone manner. After training, the network is tested on another 1,000-sample dataset.} Clearly, the network under complete labeling outperforms that under missing labeling by around 20\% in terms of both precision and recall. Moreover, it is evident that the performance of the DNN under missing labeling experiences a downward trend after the 20-th epoch, confirming the negative effects of the wrong gradient signals introduced by missing labeling. 

\subsection{Heterogeneous Modality across AVs} \label{ssec:modality_hetero}

\begin{figure}[t]
    \setlength\abovecaptionskip{6pt}
    \vspace{-2.5ex}
	   \captionsetup[subfigure]{justification=centering}
		\centering
		\subfloat[Average precision.]{
		  \begin{minipage}[b]{0.47\linewidth}
		        \centering
			    \includegraphics[width = 0.96\textwidth]{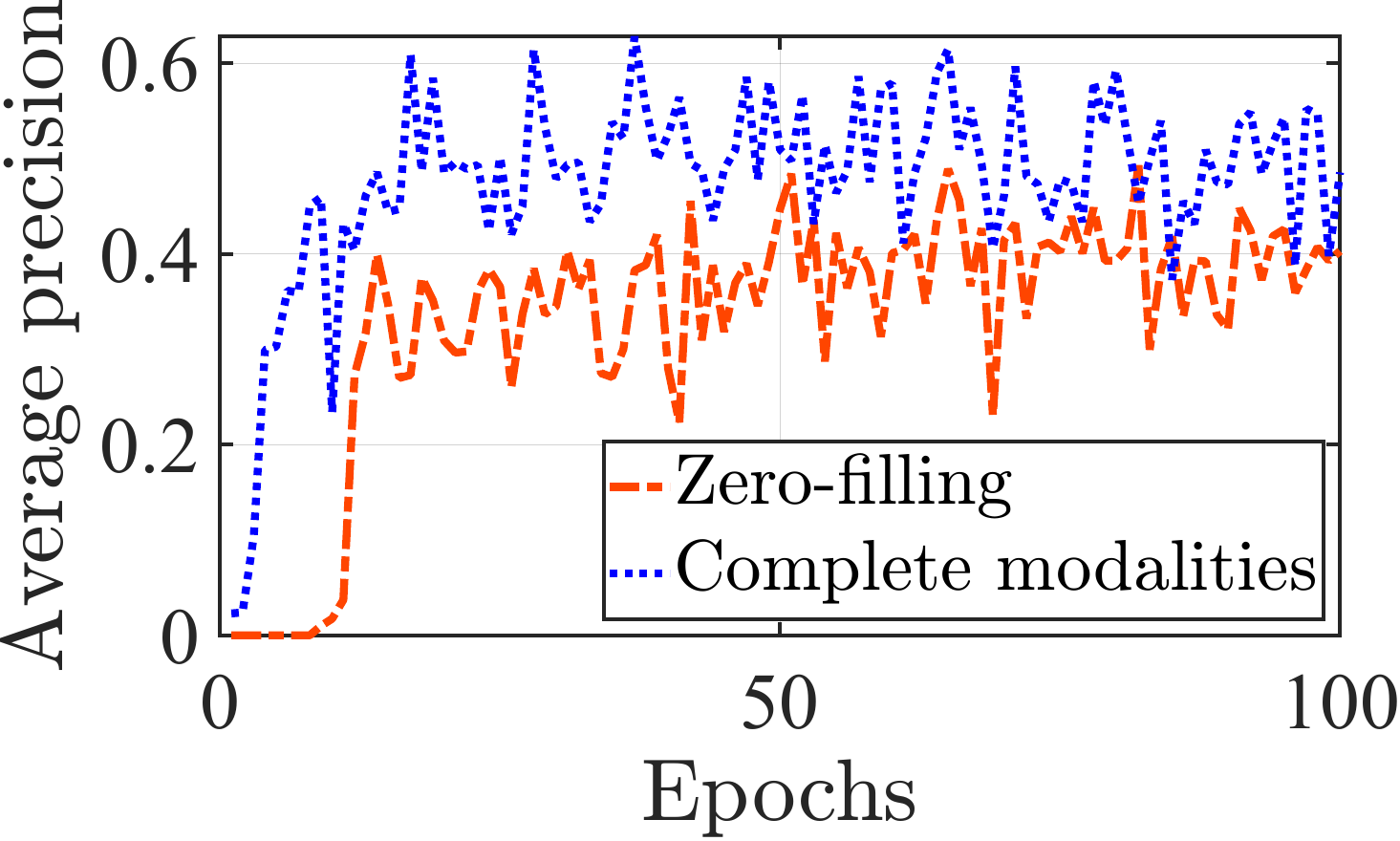}
			\end{minipage}
		}
		\subfloat[Average recall.]{
		    \begin{minipage}[b]{0.47\linewidth}
		        \centering
			    \includegraphics[width = 0.96\textwidth]{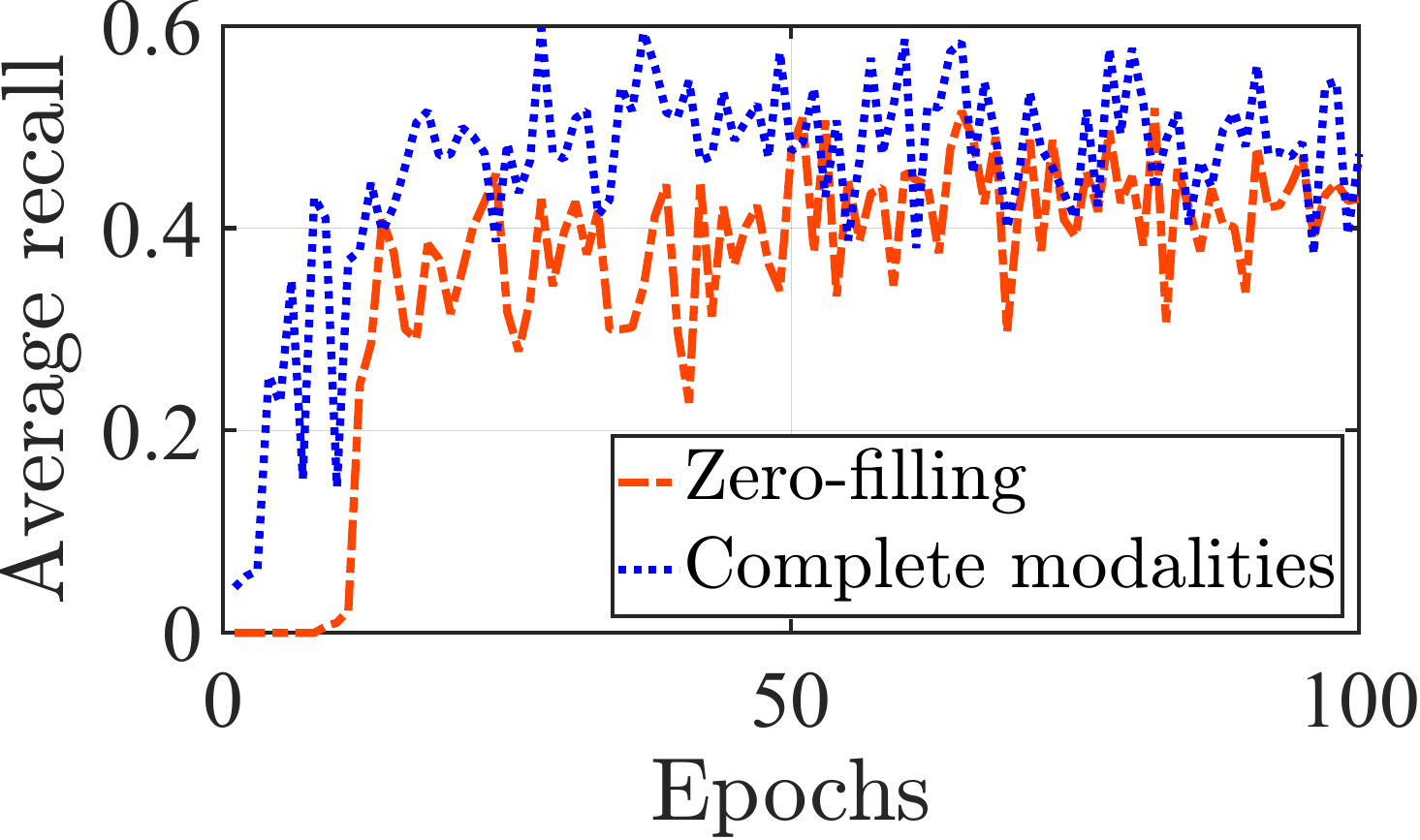}
			\end{minipage}
		}
		\caption{Damaging effects of missing modality.}
		\label{fig:damage_miss_mod}
	    \vspace{-2ex}
\end{figure}

Most prior work on the fusion of multimodal sensing data assumes that all modalities are available for every training data point~\cite{xu2018pointfusion, bijelic2020seeing}. This assumption may not be valid in reality, as the sensing modalities of different AVs are often heterogeneous for two reasons. On one hand, the AVs may be equipped with different types of sensors by their manufacturers. On the other hand, even for AVs from the same manufacturer, it is common that certain sensor experiences malfunctions, causing corresponding data modality to get lost. Such heterogeneous modalities  pose significant challenges to DNN-based OD. 
Removing data entries with missing modalities or keeping only modalities shared among all clients can be a makeshift, but useful information conveyed in other modalities or clients can be discarded. \brev{Lacking access to global statistics also renders filling a missing modality with typical statistics (e.g., mean) impractical, leaving zero-filling~\cite{van2018flexible} as the only possibility. Therefore, we show the average precision and recall of an OD network in Figure~\ref{fig:damage_miss_mod}; the model is trained in a standalone manner under complete modalities and missing modalities with zero-filling. In the training process, data with missing radar and lidar each accounting for 25\% of a 1,000-sample dataset.} The results demonstrate that the precision and recall of training the models with complete modalities outperform those with partial modalities by more than 20\% and 10\%, respectively, confirming that zero-filling does not fully overcome the challenge. To mitigate the missing modality, it is necessary to propose a new data imputation technique.

\subsection{Model Divergence}\label{ssec:environ_hetero}

\begin{figure}[h]
    \setlength\abovecaptionskip{6pt}
    \vspace{-2.5ex}
	   \captionsetup[subfigure]{justification=centering}
		\centering
		\subfloat[PCA embedding.]{
		  \begin{minipage}[b]{0.47\linewidth}
		        \centering
			    \includegraphics[width = 0.96\textwidth]{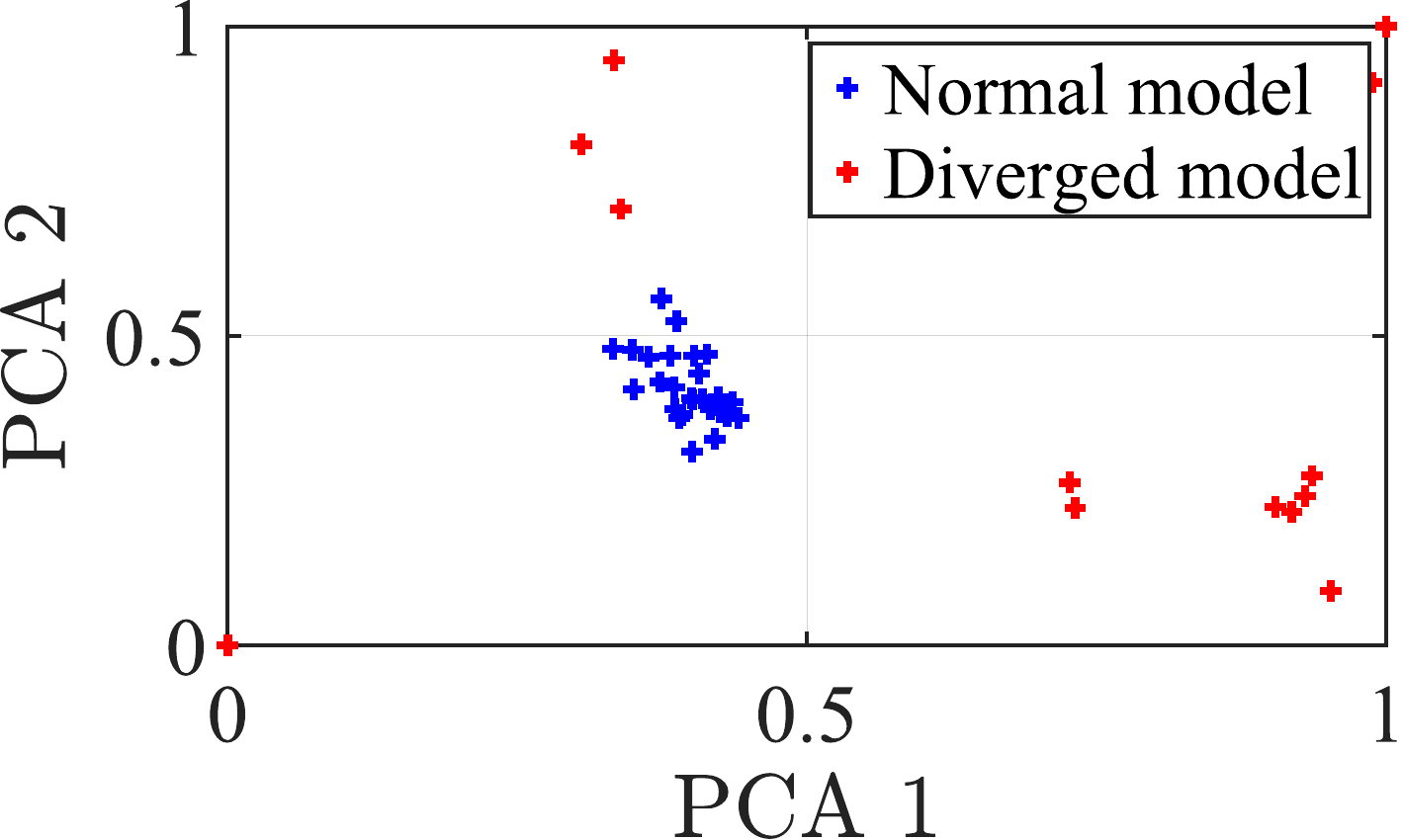}
			    \label{subfig:pca_diverging}
			\end{minipage}
		}
		\subfloat[Average precision.]{
		    \begin{minipage}[b]{0.47\linewidth}
		        \centering
			    \includegraphics[width = 0.96\textwidth]{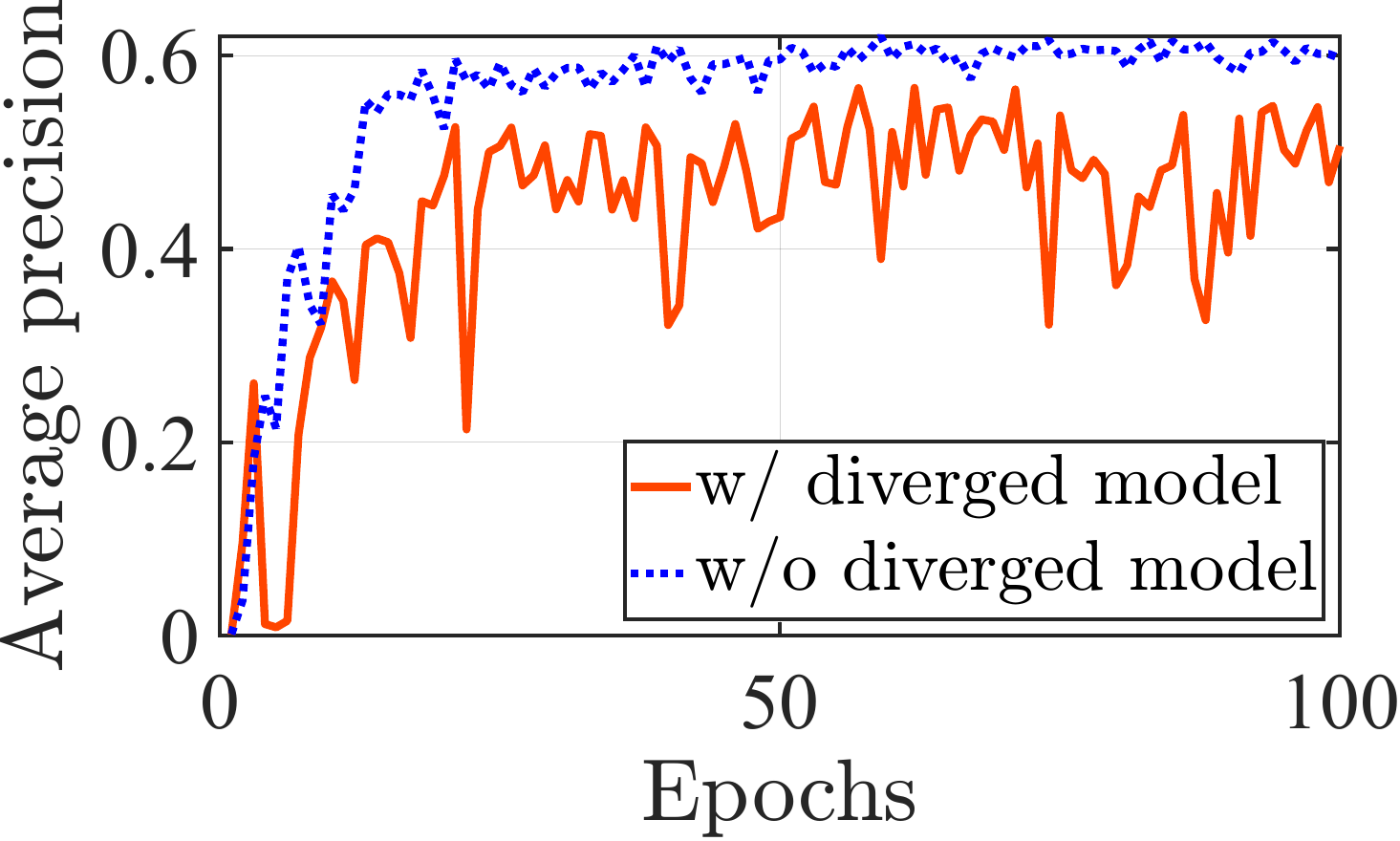}
			    \label{subfig:pre_diverging}
			\end{minipage}
		}
		\caption{Damaging effects of diverging models.}
		\label{fig:div_gradient}
	    \vspace{-2ex}
\end{figure}

Besides the above label and modality heterogeneity across the clients, there exist other heterogeneities such as those introduced by environments (e.g., different weather and road conditions). Such heterogeneity makes the local models on AVs to be diverged and the optimization goal can even become contradictory. %
We demonstrate such model divergence in Figure~\ref{subfig:pca_diverging}, \brev{where we involve 40 clients each holding a 1,000-sample dataset for training. These datasets have i) annotation level ranging from 10\% to 100\% (with 10\% step size for every 4 clients), ii) 25\% chance to hold data with missing radar or lidar modality, and iii) equal chance to have data recorded under clear, foggy, rainy, and snowy weather. After training, the network is tested on another 2,000-sample dataset.} \brev{We apply} PCA (principal component analysis)~\cite{pearson1901liii} to the model weights \brev{at the 10-th epoch, }%
and visualize the first two PCA components, one may readily observe that, while about half of local model weights form a cluster (colored in blue), there exist multiple outliers (colored in red). If we recklessly perform aggregation on these model weights, the performance of federated model will be significantly degraded by the outliers. We demonstrate the effects of diverged models in Figure~\ref{subfig:pre_diverging}. The results show that aggregating the diverged models leads to a 10\% drop in OD precision when compared with the aggregated model from homogeneous training data.

%% file: 3_design.tex
\section{System Design}\label{sec:system_design}
Based on our discussions in \S~\ref{sec:background_motivation}, we hereby present \sysname\ comprising a two-level design: i) \brev{a multimodal OD} network
to fully exploit the information provided by multimodal sensors equipped on the AVs,
and ii) an FL framework involving specifically designed loss, missing modality completion module, and client selection mechanism, aiming to achieve heterogeneity-aware federated multimodal OD on distributed AVs. In the following, we first define our problem concretely, and then we present the details of the multimodal OD network and FL framework.

\subsection{Problem Statement and Overview}\label{ssec:statement}

The ultimate goal of \sysname\ is to make use of the crowdsensed data collected from multiple AVs (i.e., clients) to increase the data diversity, thus improving upon the performance of a standalone OD network deployed on a single client. Since the sensors 
on AVs can have multiple viewing perspectives, i.e., the lidar, radar, and camera have 3D, bird's-eye view, and front view, respectively, there are no one-size-fits-all solutions. Therefore, we specifically choose to solve the \textit{vehicle detection} problem~\cite{petrovskaya2009model} (a special case of OD) from the bird's-eye view \brev{using lidar and radar}, thanks to (also confined by) the availability of dataset and vehicle annotations~\cite{oxford_robotcar}. \brev{We avoid using camera in \sysname\ for two reasons. First, the perception capability of lidar and camera largely overlap due to their similar spectrums. Second, the current settings and network architecture mostly are focused on the bird’s eye view of the vehicle’s surroundings, making the camera’s orthogonal front view largely incompatible. }Note that our \sysname\ framework is not limited to any specific OD tasks, because performing vehicle detection actually encompasses all critical elements in fulfilling other OD tasks.

Since AV scenarios are by default a distributed setting, FL is a good candidate for better utilizing the data diversity from geographically distributed clients. \brev{However, combining FL with OD may exacerbate OD's chaotic loss surface emphasized in \S~1, 
forcing naive aggregation algorithms to yield only comparable or even inferior performance compared to traditional standalone training~\cite{liu2020fedvision, jallepalli2021federated, yu2019federated}, especially under the challenges mentioned in \S~\ref{sec:background_motivation}.
Fortunately, our insight indicates that high tolerance to data anomalies can often allow effective training that leads to meaningful local minimums, by smoothly navigating on the chaotic loss surface; this motivates our following design considerations. Firstly, both data preprocessing and network architecture should be modularized and flexible enough to accommodate potentially abnormal multimodal inputs. Secondly, the network should be equipped with a mechanism to tolerate annotation anomalies of the input data. Thirdly, there should be a way to fill in missing modalities without making the data distributions abnormal. Finally, the aggregation mechanism must be sufficiently robust to withstand potentially diverging client models that could result in a non-optimal outcome after aggregation.}

\subsection{Multimodal Vehicle Detection} \label{ssec:multimodal_network}

Before introducing the \sysname\ framework, we first look at how to design a multimodal OD network under an FL setting. While the design method of a conventional two-stage OD network is well-established, 
how to integrate multimodal processing into the network remains an open issue. Furthermore, extending the multimodal OD network to the FL scenario put more stringent requirements on handling of the multimodal data. Intuitively speaking, different data modalities of such a network should i) conform to similar data formats, thus facilitating multimodal fusion, ii) collaborate by sharing information so as to enhance other modalities, and iii) be loosely coupled so as to support a more flexible FL pipeline that better deals with heterogeneous data and environment. In this section, we start with introducing the basics of the conventional OD network. Then we align lidar and radar data for improving data compatibility, in order to satisfy the requirement i). Finally, we present a novel feature-level fusion technique to satisfy the other two requirements with strong information sharing and loose coupling among the modalities. 

\subsubsection{Object Detection Basics} \label{sssec:od_basics}

\begin{figure}[h]
    \setlength\abovecaptionskip{6pt}
	\centering
	\includegraphics[width=0.95\linewidth]{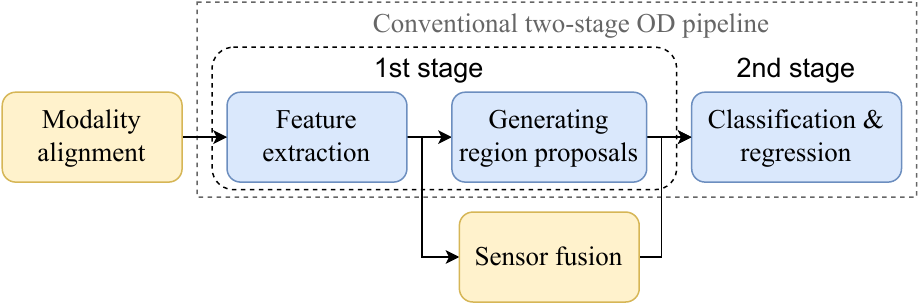}
	\caption{The upgraded OD pipeline of \sysname's multi-model vehicle detection network.}
	\label{fig:object_detection}
\end{figure}

Conventional two-stage OD follows 3 major steps~\cite{girshick2015fast, ren2015faster}, with 2 steps in the first stage as shown by the ``blue'' boxes in Figure~\ref{fig:object_detection}. Generally, a feature map is first extracted using well-accepted feature extractors (e.g., VGGNet~\cite{simonyan2014very} or ResNet~\cite{he2016deep}), then region proposals are generated by the region proposal network (RPN). Specifically, taking the feature maps as input,
RPN generates anchor boxes with pre-defined fixed scales and aspect ratios. The built-in \newrev{classifier of RPN differentiates} whether each anchor box is foreground or background. \newrev{The outcome allows RPN to generate the region proposals; it leverages a built-in regressor to fit the anchor boxes to their corresponding objects by adjusting their offsets. With the above completing Step-1, Step-2 involves the region proposals being filtered by non-maximum suppression (NMS): the proposals with the highest confidence are selected and excessive proposals overlapping with higher-confidence proposals above a given threshold are removed.} The loss of RPN is $L^{\mathrm{RPN}} = L_{\mathrm{cls}}^{\mathrm{RPN}} +  L_{\mathrm{loc}}^{\mathrm{RPN}}$, where $L_{\mathrm{cls}}^{\mathrm{RPN}}$ is a binary cross entropy (BCE) loss \newrev{measuring} the ``objectness'' of the classification (i.e., how good the RPN is at labelling the anchor boxes as foreground or background), and $L_{\mathrm{loc}}^{\mathrm{RPN}}$ is an $L^1$ loss \newrev{quantifying} the localization performance of the predicted regions generated by the RPN.

\newrev{The second stage (also Step-3) performs a fine-tuning to} jointly optimizes a classifier and bounding-box regressors. After cropping out the feature map and RoI pooling~\cite{ren2015faster} of the interested region according to the generated proposals, it further uses a classifier to detect whether the generated bounding box contains a specific class of object. It also fine-tunes the bounding boxes using a class-specific regressor. \newrev{Essentially, this stage} introduces three losses, i.e., a BCE classification loss $L_{\mathrm{cls}}$ measuring \newrev{the network performance in} labeling a predicted box with an object, an $L^1$ box regression loss $L_{\mathrm{reg}}$ \newrev{quantifying} how the predicted location deviates from the true location, and a BCE direction loss $L_{\mathrm{dir}}$ \newrev{specifying} whether the vehicle is pointing upward or downward to remove ambiguity, thus confining the possible angles of the rotated bounding box to be in the range of $[0^{\circ}, 180^{\circ}]$. In summary, the \newrev{overall} loss function for the OD network can be written as:
\begin{align}
    L_{\mathrm{total}} = L^{\mathrm{RPN}} +  L_{\mathrm{cls}} +  L_{\mathrm{reg}} + L_{\mathrm{dir}}
\end{align}

\begin{figure*}[t]
    \setlength\abovecaptionskip{8pt}
	\centering
	\includegraphics[width=0.92\linewidth]{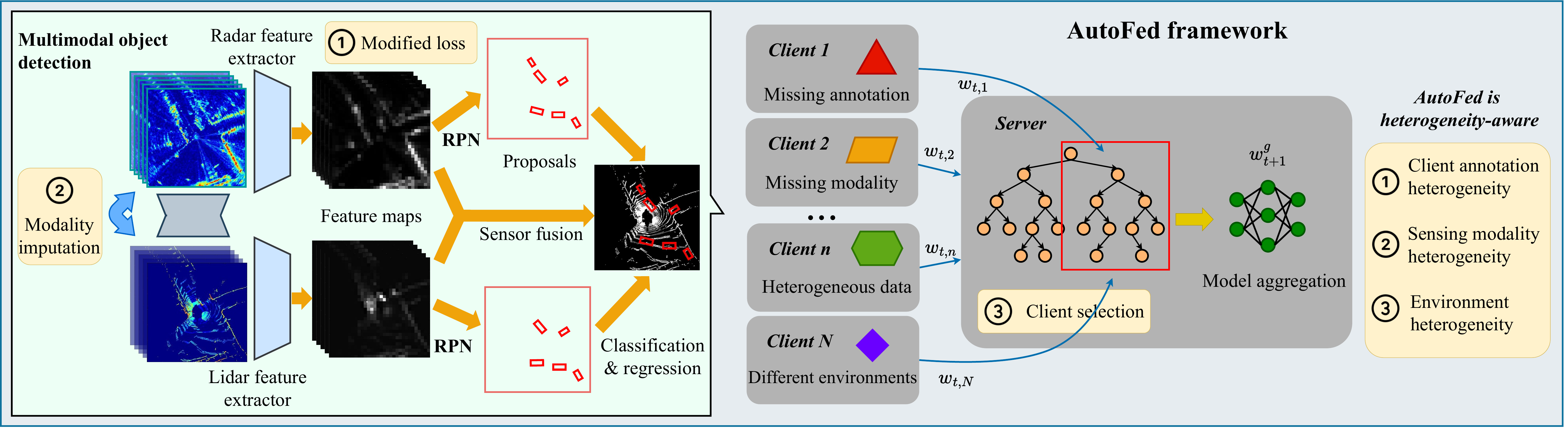}
	\caption{\sysname\ architecture: Federated multimodal learning with heterogeneity-awareness.}
	\label{fig:arch}
\end{figure*}

\subsubsection{Modality Alignment} \label{sssec:mod_alignment}
The heterogeneous data generated by multiple modalities poses a challenge to conventional OD network. Specifically, the input 3-D lidar point clouds and mechanically scanned 2-D radar heatmap are incompatible and cannot be readily fused or imputed (as will be explained in \S~\ref{sssec:imputation}) on both the original data space and the feature space. To reconcile the incompatibility, we first voxelize the 3-D point cloud obtained by lidar~\cite{yang2018pixor}. Since we are interested in performing vehicle detection from the bird's-eye view, the horizontal 2-D slices of the resulting lidar data can be deemed as an image with 36 channels, i.e., 35 channels depicting the point occupancy in the space and 1 channel \newrev{indicating} the overall intensity of the lidar signals obtained on the horizontal plane. Similarly, the radar signal can be deemed as an image with a single channel since it has no 3-D information. After converting the data into ``multi-channel'' images, they are further registered by considering the extrinsics and resolutions of the sensors, as well as vehicle kinematics. \newrev{Finally, two independent yet identical feature extractors (those of the OD network as shown on the left side of Figure~\ref{fig:object_detection}) are used} to process lidar image $\mathbf{x}_l \in \mathcal{L}$ and radar image $\mathbf{x}_r \in \mathcal{R}$, where $\mathcal{L}$ and $\mathcal{R}$ are the datasets containing lidar and radar images, respectively. \newrev{While the same architecture of these feature extractors guarantees that the modality alignment is preserved on the feature space, 
they differ in the number of input channels to} cater the respective needs of lidar and radar data.

\subsubsection{Feature-Level Sensor Fusion} \label{sssec:fl_fusion}
Two approaches exist for fusing multimodal data, i.e., data-level and feature-level fusion. \sysname\ opts for feature-level fusion \newrev{thanks to its} better flexibility and low coupling offered by fusion at a \newrev{later} stage in the network. Specifically, to extend the OD network in \S~\ref{sssec:od_basics} to a multimodal setting, we further add parallel feature extractors for other modalities. Suppose the feature extractors output lidar feature map $\mathbf{z}_l$ and radar feature map $\mathbf{z}_r$, one naive method to perform feature-level fusion would be to concatenate $\mathbf{z}_l$ and $\mathbf{z}_r$, and feed the concatenated features to \newrev{Step-2 of the OD} network. However, this straightforward method fails to exploit the inter-modality relationship. A more relevant approach for exploiting the relationship is to apply the cross-attention mechanism~\cite{wei2020multi}. It generates an attention mask, in which information from a different modality is harnessed to enhance the latent features of the interested modality (e.g., an attention mask derived from lidar is used to enhance radar features, and vice versa). Different from the existing self-attention mechanism~\cite{vaswani2017attention}, our cross-attention mechanism focuses on modeling the cross-correlation among different modalities, and it adaptively learns the spatial correspondence to derive better alignment of important details from different modalities.

Essentially, \newrev{our} cross-attention mechanism can be described as transforming the latent representation $\mathbf{z}$ to a query $\mathbf{q}$ and a set of key-value pair $\mathbf{k}$ and $\mathbf{v}$, and then mapping them to an output. The query, keys,  and values are all linearly transformed versions of the \newrev{input $\mathbf{z}_s\!: s \in \{\text{lidar}, \text{radar}\}$:}
\begin{align}\label{eq:qkv}
\mathbf{q}_s = \mathbf{W}_q\mathbf{z}_{\bar{s}} + \mathbf{b}_q,~~~
\mathbf{k}_s = \mathbf{W}_k \mathbf{z}_{\bar{s}} + \mathbf{b}_k,~~~
\mathbf{v}_s = \mathbf{W}_v \mathbf{z}_s + \mathbf{b}_v,
\end{align}
where $\bar{s}$ is the complementary sensing modality of $s$ (e.g., if $s$ is radar, then $\bar{s}$ is lidar, and vice versa), $\mathbf{W}_q$, $\mathbf{W}_k$, $\mathbf{W}_v$ and $\mathbf{b}_q$, $\mathbf{b}_k$, $\mathbf{b}_v$ are trainable matrices and vectors that help transforming the input to its corresponding query $\mathbf{q}_s$, key $\mathbf{k}_s$, and value $\mathbf{v}_s$, whose dimensions are denoted by $d_q$, $d_k$, $d_v$, respectively. The output context $\mathbf{z^{\prime}}_s$ is obtained as a weighted sum of the values in $\mathbf{v}_s$, where the weight of each value is a normalized product of the query $\mathbf{q}_s$ and its corresponding key $\mathbf{k}_s$: 
$
\mathbf{z^{\prime}}_s=\operatorname{softmax}\left(\frac{1}{\sqrt{d_k}}\mathbf{q}_s \mathbf{k}_s^{T} \right) \mathbf{v}_s.
$

\subsection{\sysname\ Framework} \label{ssec:autofed_sysdesign}
We intend to design an FL framework that extends \newrev{our} multimodal vehicle detection network in \S~\ref{ssec:multimodal_network} to a training scenario where the data are collected by geographically distributed AVs. As illustrated in Figure~\ref{fig:arch}, \sysname\ improves the multimodal vehicle detection network in three aspects: i) modifying the loss of RPN to deal with client annotation heterogeneity, ii) employing an autoencoder to perform data imputation of missing sensing modalities, and iii) applying a client selection strategy based on $k$-d tree~\cite{bentley1975multidimensional} to overcome the diverged models brought by the environment \newrev{and aforementioned} heterogeneity.

\subsubsection{\brev{Modified Loss Function for Tolerating Annotation Anomalies}}\label{sssec:mce}
As stated in \S~\ref{ssec:label_skew}, the heterogeneity of labeled data may send wrong gradient signals during \sysname\ training, since the bounding boxes that should be classified as foreground otherwise \newrev{can be} wrongly labeled as background when their correct annotations are missing. The motivation for our modified loss is that, despite the lack of correct annotations, the \sysname\ model can identify vehicles wrongly labeled as backgrounds according to its own well-established classifier, \brev{thus avoiding sending erroneous gradient signals during backpropagation and better guiding the convergence
on the OD loss surface mentioned in \S~\ref{ssec:statement}.} Specifically, if the feature map of an anchor region is found to be similar to a vehicle, the classifier naturally assigns a high probability $p$ of predicting it as a vehicle. \brev{This comes under a reasonable assumption that, since the global model is trained sufficiently with on average high-quality annotations, it can be more trustworthy than the annotations from a few incompetent clients.} %
Recall \newrev{the BCE loss of RPN in \S~\ref{sssec:od_basics} as: $L_{\mathrm{cls}}^{\mathrm{RPN}} = -p^{*} \log \left(p\right)-\left(1-p^{*}\right) \log \left(1-p\right)$, where $p^{*}$ is the training label taking on values of 0 or 1, \brev{respectively indicating the bounded region being background or vehicle}. 
Consequently, the modified cross-entropy (MCE) loss becomes:}
\begin{align}
\left\{\begin{matrix}
0, & p>p_{\mathrm{th}}~\text{and}~p^{*}=0, \\ -p^{*} \log p-(1-p^{*}) \log (1-p), & \text{otherwise},
\end{matrix}\right.
\end{align}
where $p_{\mathrm{th}}$ is a threshold value after which we believe that the classifier is more trustworthy than the annotations. \brev{The value of $p_{\mathrm{th}}$ is determined by hyperparameter search in \S~\ref{sssec:loss_thresh}.}

\begin{figure}[h]
    \setlength\abovecaptionskip{6pt}
    \vspace{-1.5ex}
	   \captionsetup[subfigure]{justification=centering}
		\centering
		\subfloat[Average precision.]{
		  \begin{minipage}[b]{0.47\linewidth}
		        \centering
			    \includegraphics[width = 0.96\textwidth]{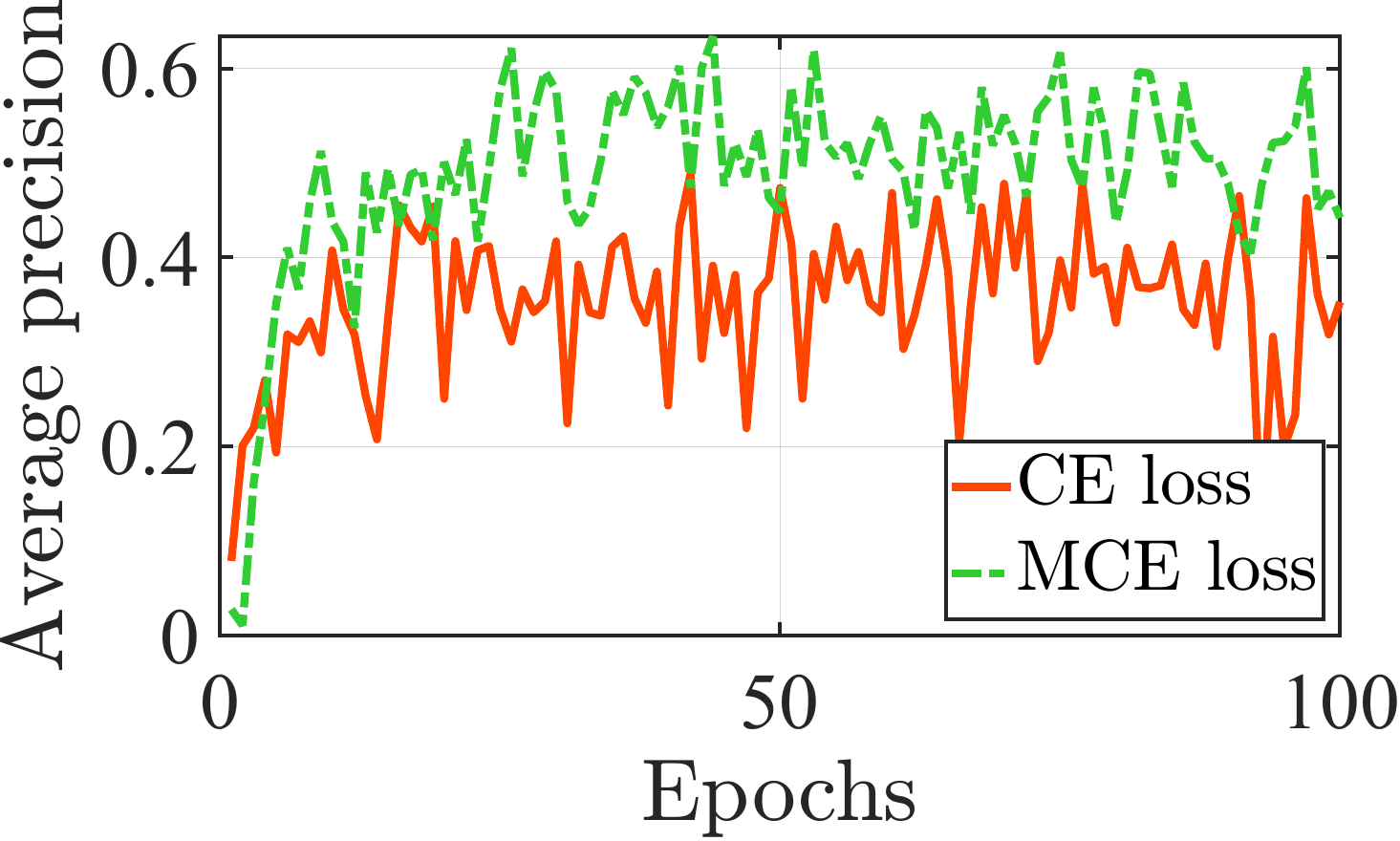}
			\end{minipage}
			\label{subfig:effect_loss_ap}
		}
		\subfloat[Average recall.]{
		    \begin{minipage}[b]{0.47\linewidth}
		        \centering
			    \includegraphics[width = 0.96\textwidth]{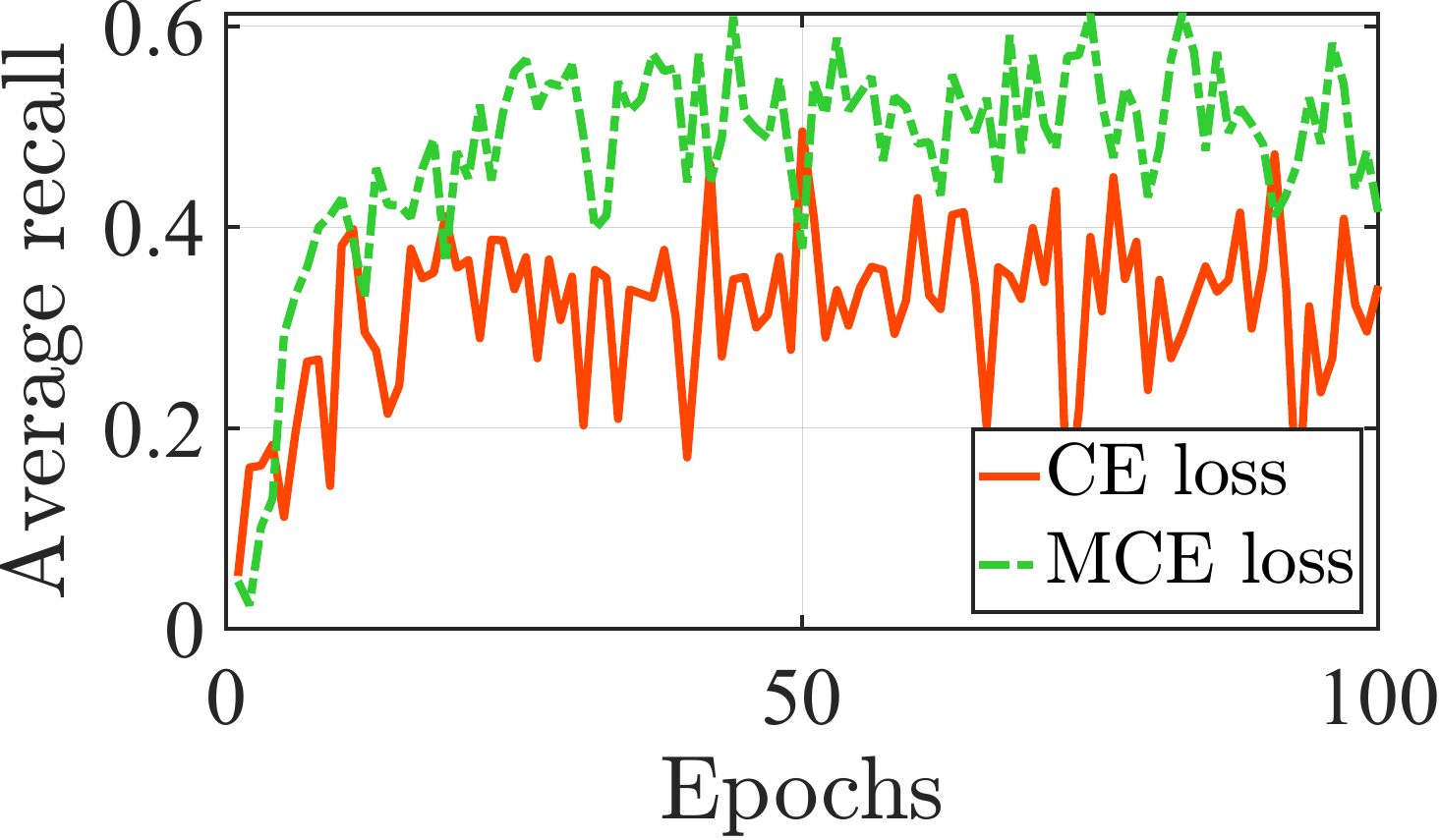}
			\end{minipage}
			\label{subfig:effect_loss_ar}
		}
		\caption{\newrev{Comparison between CE and} MCE loss.}
		\label{fig:effect_loss}
		\vspace{-.5ex}
\end{figure}

To demonstrate the efficacy of the MCE loss, we train a multimodal vehicle detection network \brev{using the settings in \S~\ref{ssec:label_skew}}. The training results of regular CE loss and our MCE are shown in Figure~\ref{fig:effect_loss}; \newrev{they evidently confirm the superiority of MCE loss,
though the average precision and average recall of both CE and BCE losses} fluctuate around their means after sufficient training (approximately 15 epochs). \newrev{First of all}, The average precision of vehicle detection is respectively 0.57 and 0.4 when CE and MCE loss are used. Similarly, \newrev{there is a gap greater than 0.1}  in the average recalls when the two losses are used. Moreover, it is \newrev{clear that, though} training with CE loss achieves higher precision and recall in the few initial epochs, it is quickly overtaken by the MCE loss, which keeps an upward trend and converges faster. Last but not least, one may also observe that there is a slight downward trend of performance when the CE loss is used after the 15-th epochs. The performance gaps and different performance trends clearly demonstrate that our MCE loss can make full use of vehicle annotations while avoiding backpropagating erroneous gradients caused by missing annotations.

\subsubsection{\brev{Modality Imputation with Autoencoder for Tolerating Modality Anomalies}} \label{sssec:imputation}

We have shown in \S~\ref{ssec:modality_hetero} that conventional data imputation methods (e.g., filling the missing modalities with 0’s) incurs information loss, and may even introduce biases into the network. To leverage the valuable information in the heterogeneous sensing modalities, we propose to fill in the missing data by leveraging the relations among different modalities. Since different modalities are aligned and loosely coupled (as explained \S~\ref{sssec:mod_alignment} and \S~\ref{sssec:fl_fusion}), we employ a convolutional autoencoder with residual connections (which connects a layer to further layers by skipping some layers in between, thus facilitating information flow) to directly perform modality imputation. The encoder of the autoencoder consists of 4 convolutional layers, and correspondingly, the decoder of the autoencoder consists of 4 transposed convolutional layers. \brev{Consequently, the lightweight architecture of our autoencoder only incurs negligible overhead representing an increase of only 4.38\% (3.129~\!GFLOPS vs. 2.988~\!GFLOPS) from the \sysname\ variant without autoencoder.} It should be noted that the autoencoder is pre-trained and does not participate in the training process of \sysname. During the pretraining stage, the autoencoder aims to learn a latent representation, and reconstruct the missing modality. For example, when the radar modality $\mathcal{R}$ is missing, the autoencoder encodes the lidar modality $\mathcal{L}$ and translates the latent information therein to fill in the missing radar modality. 

\begin{figure}[h]
    \setlength\abovecaptionskip{6pt}
    \vspace{-1.5ex}
	   \captionsetup[subfigure]{justification=centering}
		\centering
		\subfloat[Average precision.]{
		  \begin{minipage}[b]{0.47\linewidth}
		        \centering
			    \includegraphics[width = 0.96\textwidth]{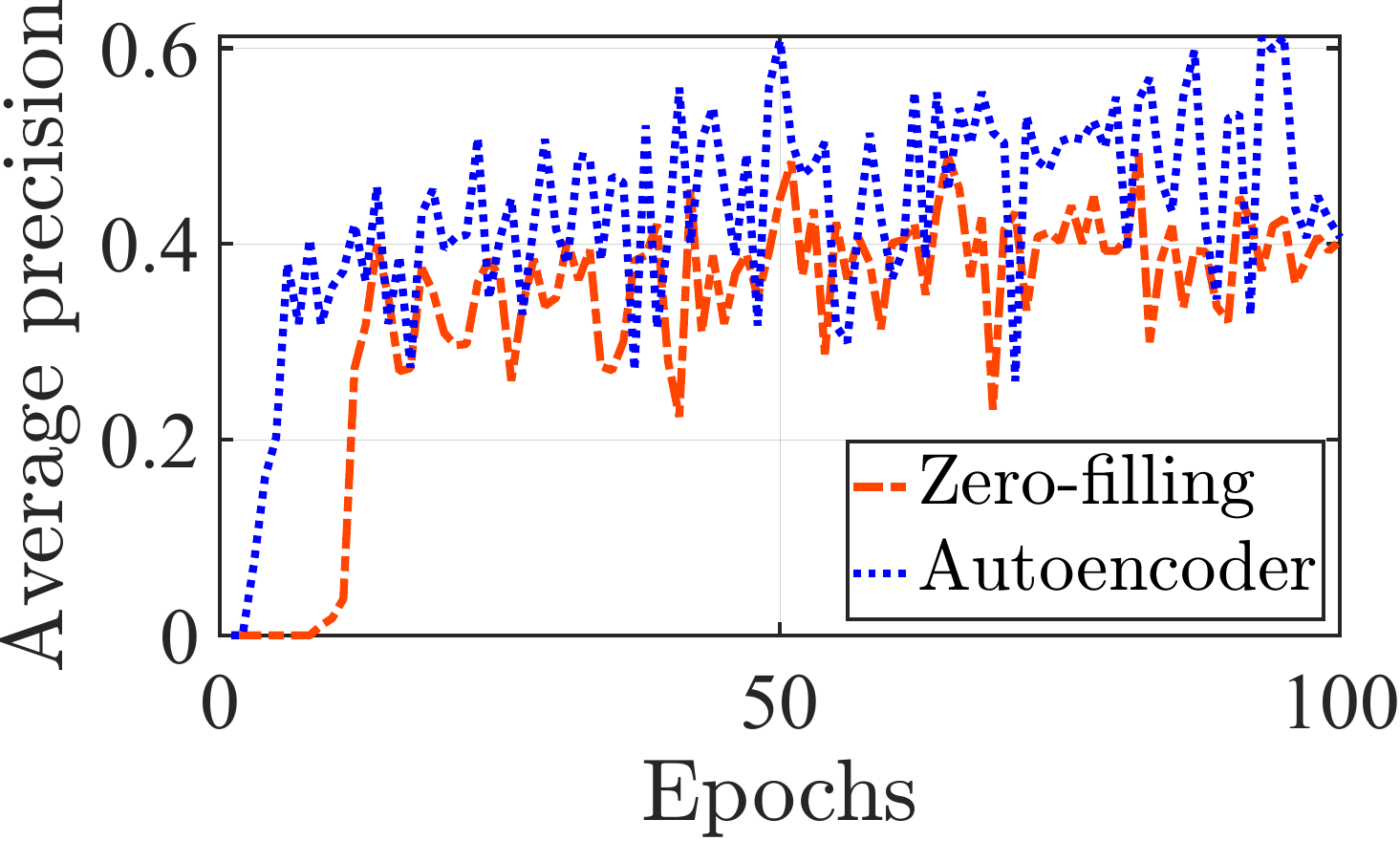}
			\end{minipage}
			\label{subfig:effect_ae_ap}
		}
		\subfloat[Average recall.]{
		    \begin{minipage}[b]{0.47\linewidth}
		        \centering
			    \includegraphics[width = 0.96\textwidth]{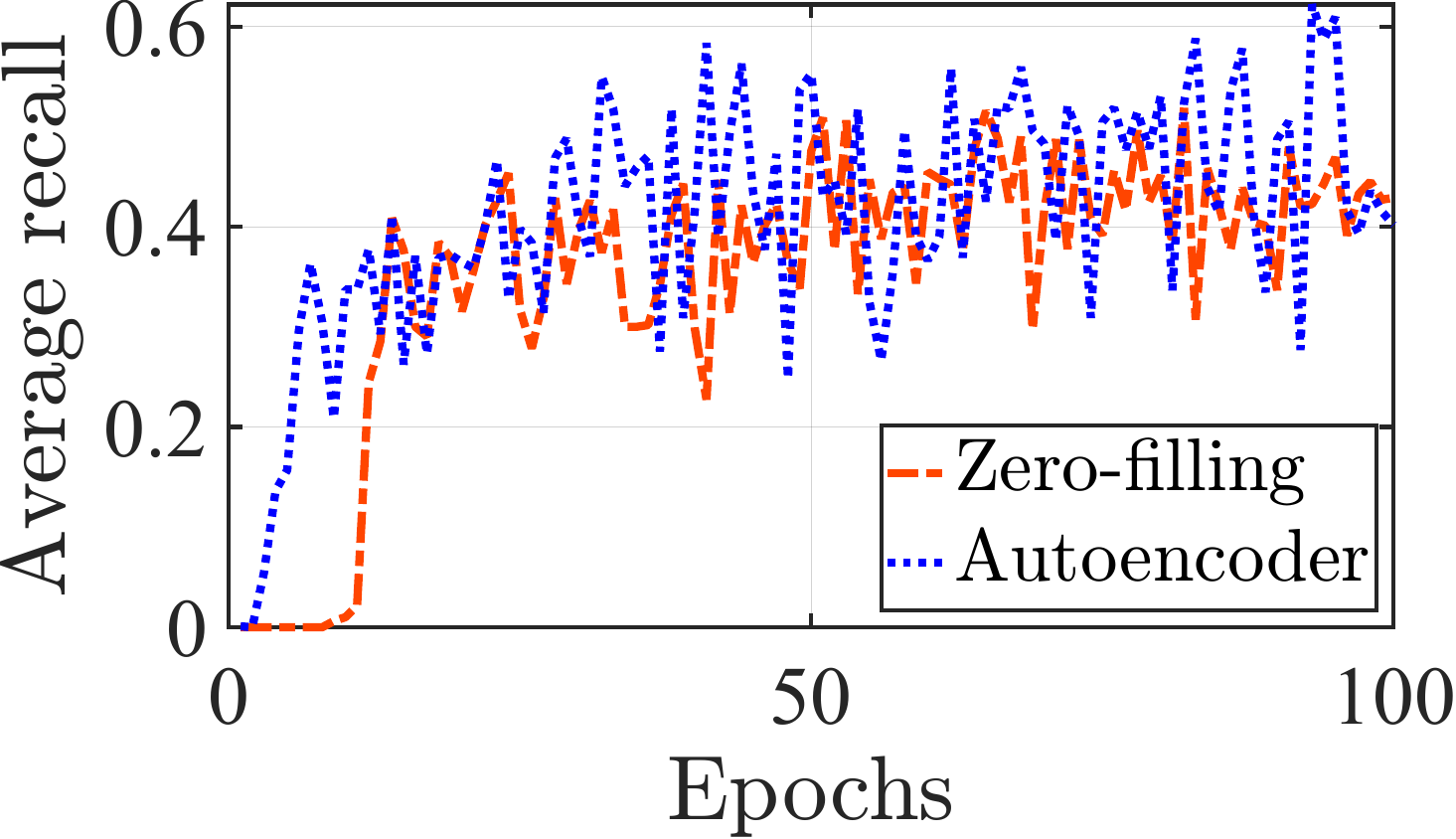}
			\end{minipage}
			\label{subfig:effect_ae_ar}
		}
		\caption{Modality imputation with an autoencoder.}
		\label{fig:effect_ae}
		\vspace{-.5ex}
\end{figure}

To show the efficacy of the above method, we train the multimodal vehicle detection network \brev{following the settings in \S~\ref{ssec:modality_hetero}}, and compare the average precision and recall of autoencoder imputation with zero-filling in Figure~\ref{subfig:effect_ae_ap} and~\ref{subfig:effect_ae_ar}, respectively. One may readily observe that \newrev{zero-filling only achieves an average precision of approximately 0.4, lower than an average precision of about 0.5 achieved by our autoencoder imputation.} Similarly, autoencoder imputation also surpasses zero-filling in terms of average recall by a \newrev{discernible} margin. \newrev{Figure~\ref{fig:effect_ae} also indicates 
that autoencoder imputation only takes about 5 epochs to converge, much faster than} the convergence speeds (i.e., 10 and 15 epochs) by zero-filling. The higher average precision and recall, as well as the faster convergence training speed \newrev{have clearly demonstrated} that our designed autoencoder makes full use of the heterogeneous data by taking into account the \newrev{correlations} among different modalities. 

\subsubsection{\brev{Client Selection for Tolerating Model Weight Anomalies}} \label{sssec:client_sel}
\brev{Environment heterogeneities, including different weather and road conditions (as indicated in \S~\ref{ssec:environ_hetero}), as well as other human-induced heterogeneities (e.g., inaccurate annotations), are not easily solvable using the techniques described in Sections~\ref{sssec:mce} and~\ref{sssec:imputation}, yet they can cause serious model divergence among the clients.} 
Training with \brev{diverging }%
clients holding extremely biased datasets may contradict models from other clients, thus increasing the overall losses. \brev{To make things worse, the chaotic loss surface mentioned in \S~\ref{sec:introduction} and \S~\ref{ssec:statement} can disorient the gradient descent algorithm used for training the OD model, and further diverge the model weights. }These observations urge us to devise a novel client selection strategy immune to divergence, rather than blindly using FedAvg to aggregate model weights from all clients equally. %
By selectively removing outlier clients, the client selection strategy should help the loss navigating on the surface more efficiently. %

Suppose there are $N$ clients $\{C_1, \cdots, C_n, \cdots, C_N\}$ in total, which forms a set $S$. To mitigate the issue of diverged models, we would like to dynamically select a subset $S^{\prime} =\{ C_1, \cdots, C_m, \cdots, C_M\}$ of $M$ clients ($M<N$) after each FL communication round to minimize the sum of inter-client distances of model weights. To achieve this, we propose that, after receiving the local models from the clients, the central server constructs a $k$-d tree using the received model weights. The $k$-d tree is a bisecting structure where each branch point is the median in some dimension, and this bisecting structure helps improve the efficiency of finding the \newrev{nearest client (local) models} with minimum distances. Subsequently, the central server traverses every client in the set $S$, and queries its $M-1$ nearest neighbors efficiently using the $k$-d tree data structure. The client with the minimum distance sum to its $M-1$ neighbors, together with its $M-1$ neighbors, form the subset of selected clients $S^{\prime}$. Since the time complexity of one query is $O(\log{N})$, traversing the whole set $S$ demands a complexity of $O(N\log{N})$, which saves up a lot of time when compared with $O(N^2)$ complexity of brute-force search, especially when there are many clients involved. At last, the \newrev{central server} aggregates the model weights from the selected clients in subset $S^{\prime}$, and distributes the updated global model to all clients in $S$ for training in the next communication round.

\begin{figure}[h]
    \setlength\abovecaptionskip{6pt}
    \vspace{-1.5ex}
	   \captionsetup[subfigure]{justification=centering}
		\centering
		\subfloat[PCA embedding.]{
		  \begin{minipage}[b]{0.47\linewidth}
		        \centering
			    \includegraphics[width = 0.96\textwidth]{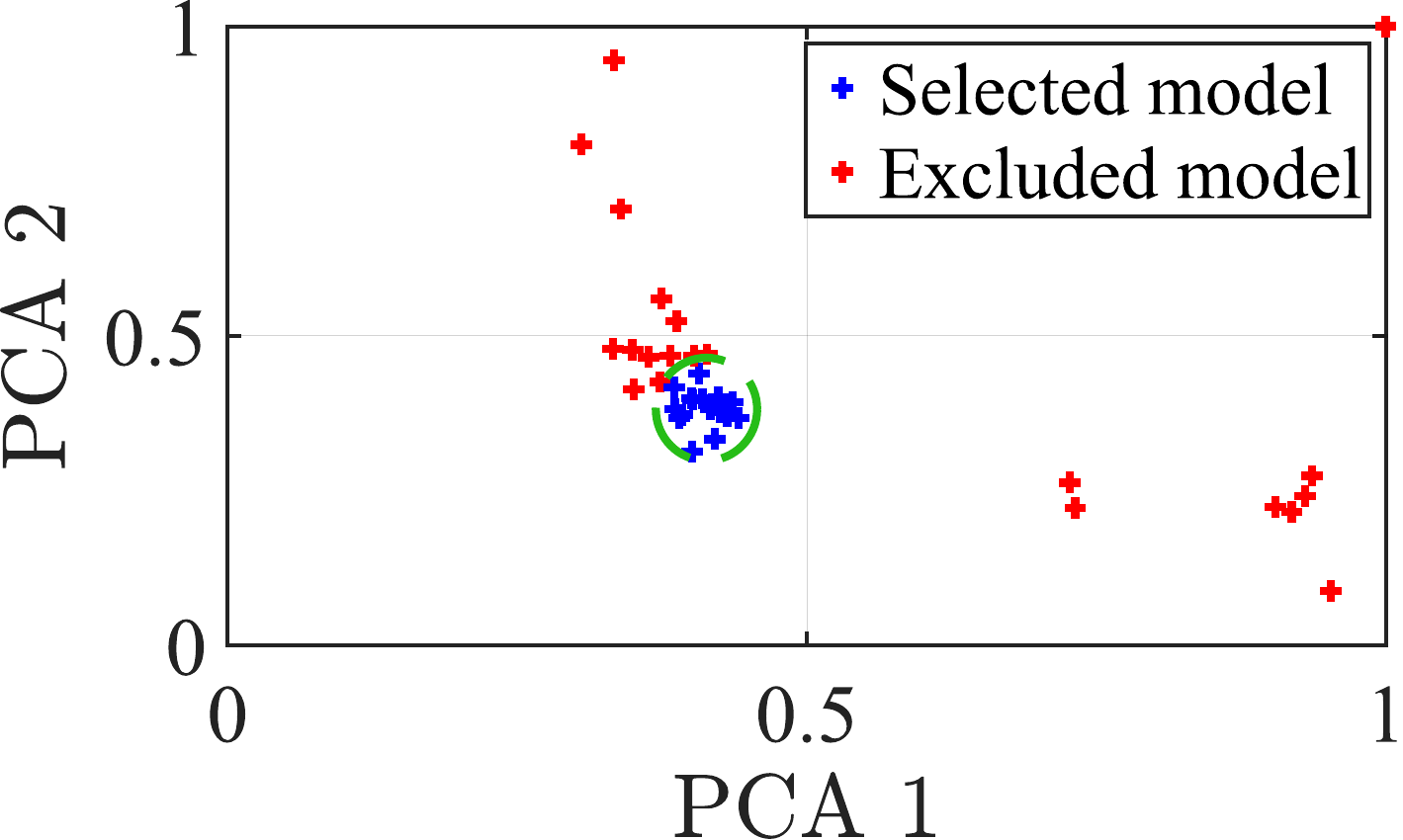}
			\end{minipage}
		    \label{subfig:client_sel_pca}
		}
		\subfloat[Average precision.]{
		    \begin{minipage}[b]{0.47\linewidth}
		        \centering
			    \includegraphics[width = 0.96\textwidth]{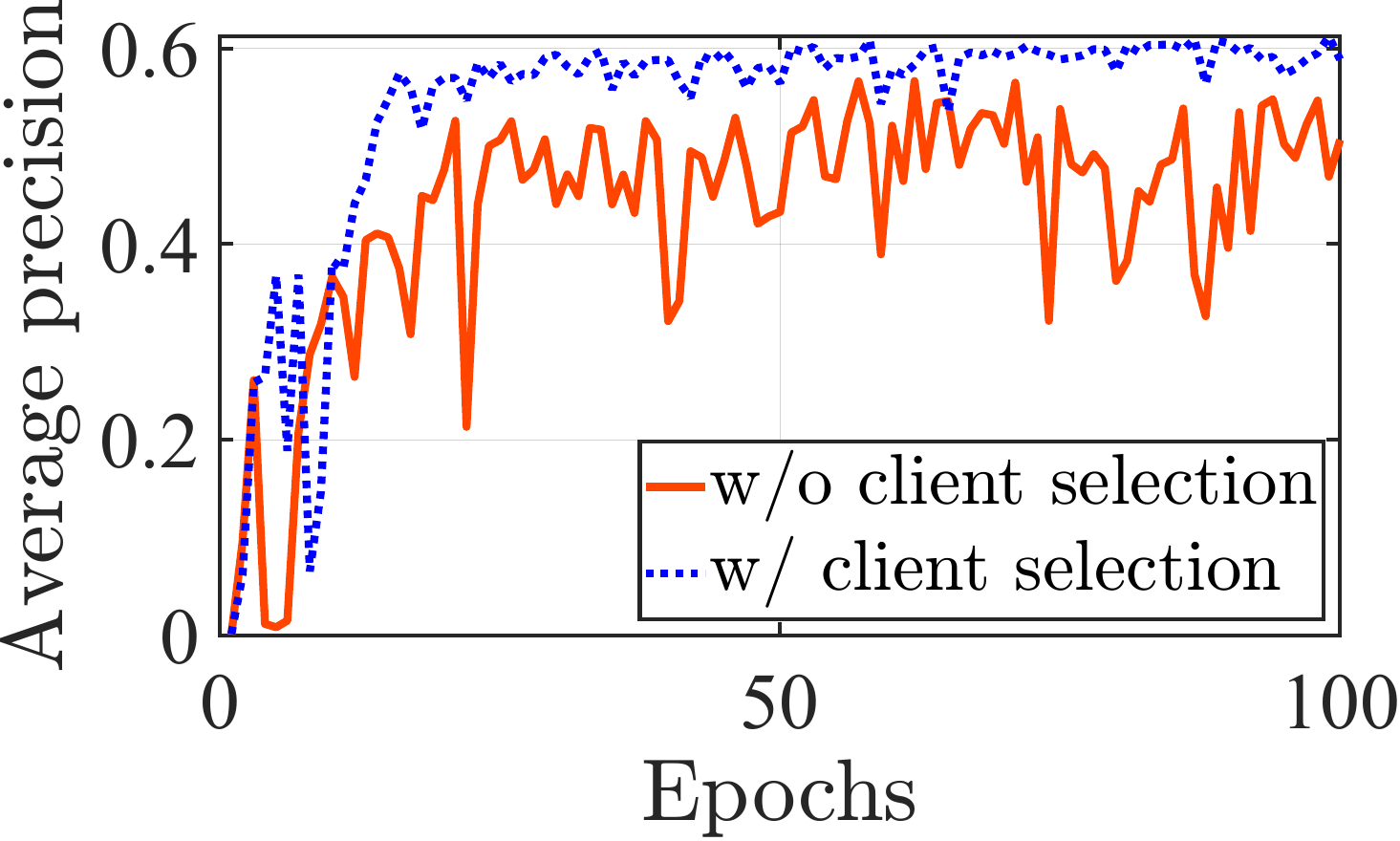}
			\end{minipage}
			\label{subfig:client_sel_precision}
		}
		\caption{Client selection mitigates diverged models.}
		\label{fig:client_sel}
\end{figure}

To illustrate the effect of our client selection strategy, we train the multi-modal vehicle detection network \brev{following the settings in \S~\ref{ssec:environ_hetero}}. After each communication round, we let the central server selects 40\% of the clients \needrev{(i.e., $M = 0.4N$)} to form a subset of clients with minimum \newrev{inter-client local model} weight distance, as demonstrated in Figure~\ref{subfig:client_sel_pca}. The average vehicle detection precision is shown in Figure~\ref{subfig:client_sel_precision}. One may readily observe that the precision of vehicle detection reaches up to 0.6 when client selection is enabled, and it fluctuates around 0.5 when model weights from all clients are aggregated using the FedAvg algorithm. Moreover, \newrev{Figure~\ref{fig:client_sel} also demonstrates} that client selection makes the training converge faster 
\newrev{with less than 20 epochs,} while the training without client selection barely starts to converge \newrev{till} the 25-th epoch. These phenomena indicate that client selection helps better utilize data from beneficial clients. Upon further inspection, we find that after convergence, the fluctuation of the precision curve with client selection is much smaller than that without client selection, which indicates that the mechanism indeed selects mutually-enhancing clients while excluding erroneous gradient signals from outliers. \brev{Additionally, it can be observed that the average precision with client selection becomes stable after only 70 epochs. This confirms that the model has effectively learned from all clients (including the corner cases), so additional training will not yield any further improvement in performance.}

\RestyleAlgo{ruled}
\LinesNumbered
\begin{algorithm}[t]
\caption{\sysname\ training.}\label{alg:fed}
\SetKwInOut{Input}{Require}
\SetKwFunction{Fns}{Server Executes}
\SetKwProg{Fn}{}{:}{}
\SetKwFunction{Fc}{Client Selection}
\SetKwFunction{Fu}{Client Update}

\Input{$N$ is the total number of clients, $c$ is the percentage of clients to choose.}
\KwData{$\{(\mathcal{L}_1, \mathcal{R}_1), \cdots, (\mathcal{L}_n, \mathcal{R}_n), \cdots, (\mathcal{L}_N, \mathcal{R}_N)\}$ where $(\mathcal{L}_n, \mathcal{R}_n)$ is the local collected lidar and radar data on the $n$-th AV.}

\Fn{\Fns}{
    initialize the global model $w^g_0$ at $t=0$\;
    $S \leftarrow \{C_1, \cdots, C_N \}$\;
    \For{communication round $t$}{
    \For{$C_n \in S$ in parallel }{
    $w_{t+1, n} \leftarrow \FuncSty{Client Update}(n)$\;
    $W_t \leftarrow W_t \cup w_{t+1, n}$\;
    }
    $M \leftarrow c\times N$\;
    $W_t^{\prime} \leftarrow \Fc(W_t, M)$\;
    $w^g_{t+1} \leftarrow \FuncSty{Model Aggregate}(W_t^{\prime})$
    }
}

\Fn{\Fu{$n$}}{
    $w_{n} \leftarrow w_t^g$ ($w_t^g$ is downloaded global model) \;
    \uIf{$\mathcal{R}_n=\varnothing$}{
    $\mathcal{R}_n \leftarrow$  \FuncSty{Radar Imputation} ($\mathcal{L}_n$)\;
    }
    \uElseIf{$\mathcal{L}_n=\varnothing$}{
    $\mathcal{L}_n \leftarrow$  \FuncSty{Lidar Imputation} ($\mathcal{R}_n$)\;
    }
    \For{each local epoch $e$}{
        \For{each batch $b$}{
            $w_{n} \leftarrow $ \FuncSty{SGD}($w_{n}, b$) \;
        }
    }
    \KwRet $w_{n}$\;
}

\Fn{\Fc{$W_t$, M}}{
    $T_t \leftarrow \FuncSty{Construct k-d Tree}(W_t)$\;
    \For{$C_i \in S$}{
        $S_i \leftarrow \FuncSty{Query k-d Tree}(T_t, C_i, M)$\;
        $d_i \leftarrow \Sigma_{m=1}^M \FuncSty{Dist}(C_i, C_m)$ \textbf{for} $C_m \in S_i$\;
    }
    $I_{min} = \argmin_i (d_i)$\;
    \For{$C_m \in S_{I_{min}}$ in parallel }{
        $W_{t, I_{min}}^{\prime} \leftarrow W_{t, I_{min}}^{\prime} \cup w_{t, m}$\;
    }
    \KwRet $W_{t, I_{min}}^{\prime}$\;
}
\end{algorithm}

\subsubsection{Putting It All Together}
We carefully summarize the training strategy of the \sysname\ framework in \textbf{Algorithm~\ref{alg:fed}}. In the algorithm, \textsf{Client Update} is the local training process for each client, \textsf{Radar Imputation} and \textsf{Lidar Imputation} are the imputation function introduced in \S~\ref{sssec:imputation}, \textsf{SGD} is the standard stochastic gradient descent algorithm with our MCE loss, \textsf{Client Selection} \newrev{has been introduced in \S~\ref{sssec:client_sel}, which includes \textsf{Construct k-d tree} and \textsf{Query k-d Tree} as the respective processes of constructing and querying $k$-d tree, as explained in \S~\ref{sssec:client_sel}, and \textsf{Model Aggregate} as} the standard process of averaging the selected local models. \brev{By putting together \sysname's modules, we create a cohesive ensemble to substantially enhance the tolerance to data anomalies. Although some techniques can be relevant even to a single model context, they work together in the FL setting to help \sysname\ navigate on the chaotic loss surface in a more robust and efficient manner. }

%% file: 4_evaluation.tex
\section{Performance Evaluation}\label{sec:evaluation}
To evaluate the performance of AutoFed, we apply AutoFed to build a vehicle detection application using the benchmark dataset~\cite{oxford_robotcar}. 
In particular, we evaluate the performance of AutoFed from four aspects: i) \newrev{comparisons with five baseline methods to demonstrate the superiority of \sysname; ii) cross-domain tests to show that \sysname\ is robust against real-life scenarios with heterogeneous data;  iii) ablation study} to show the necessity of key parameter designs, and iv) investigating the impact of FL-related hyper-parameter on the model performance. 

\subsection{Dataset}

We \brev{mainly} use the Oxford Radar RobotCar dataset~\cite{oxford_robotcar}  in our experiment. The dataset is collected by a vehicle driving around Oxford, and it includes both lidar and radar data. The lidar data is obtained by merging the point clouds collected by two Velodyne HDL-32E~\cite{velodyne} lidars \newrev{mounted on the left and right of the vehicle's top. Each} lidar sensor provides a range of 100~\!m, a range resolution of 2~\!cm, a horizontal field of view (FoV) of 360$^\circ$, and a vertical FoV of 41.3$^{\circ}$. The radar data is collected by a millimeter-wave Frequency-Modulated Continuous-Wave (FMCW) NavTech CTS350-X radar~\cite{NavTech} mounted between the two lidar sensors and at the center of the vehicle aligned to the vehicle axes. The radar achieves 2-D horizontal scan by rotation, operating with a center frequency of 76.5~\!GHz, a bandwidth of 1.5~\!GHz, a sampling rate of 4~\!Hz (hence a range resolution of 4.38~\!cm), a rotational angle resolution of 0.9$^{\circ}$, a beamwidth of 1.8$^{\circ}$, and a range up to 163~\!m; it complements lidar by providing robustness to weather conditions that may cause trouble to lidar. We further convert the data residing in the polar coordinates to Cartesian coordinates and then \newrev{calibrate radar and lidar extrinsic parameters (i.e., translation and rotation with respect to the world) by performing pose optimization to minimize} the differences between lidar and radar observations. Since there is no original ground truths for vehicle detections, we create rotated boxes by inspecting the point cloud data using Scalabel~\cite{scalabel}, which is a scalable open-source web annotation tool for various types of annotations on both images and videos.

\brev{We also involve another dataset nuScenes~\cite{caesar2020nuscenes} in our experiment to demonstrate \sysname's generalizability across datasets. The dataset contains 1 lidar and 5 radars: the lidar has 360$^\circ$ horizontal FoV, 40$^{\circ}$ vertical FoV, and 2~\!cm range resolution, while the 5 radars have 77~\!GHz center frequency and 0.1~\!km/h velocity accuracy. Unlike the radars in the Oxford dataset that perform fine-grained mechanical scans, the radars in the nuScenes dataset are fixed in positions and do not have scan capability. As a result, they only generate low-quality pointclouds. Since \sysname\ cannot demonstrate its full potential with the inferior radar modality, we limit the evaluation on the nuScenes dataset to only \S~\ref{ssec:superiority}. For both datasets, we take out a total of 50,000 samples, and use 80\% and 20\% of the total data to create training and test datasets, respectively.}

\subsection{System Implementation}
We implemented the vehicle detection application using AutoFed on multiple NVIDIA Jetson TX2~\cite{jetson} devices. The central server is equipped with an Intel Xeon Gold 6226 CPU~\cite{intel} and 128~\!GB RAM. For both \sysname\ and the baselines, we implement an FL protocol that \brev{allows 20 participating clients to randomly take 2,000 non-overlapping samples from the 40,000-sample training set.} %
Each participating client performs 5 local training epochs for \newrev{each} communication round. As for the software, Python 3.7 and PyTorch 1.9.1~\cite{pytorch} are used for implementing the application. Our vehicle detection model is built upon Detectron2~\cite{detectron2}, which is a Python library that provides state-of-the-art OD models. In particular, the settings for the multimodal vehicle detection model are as follows:
\begin{itemize}
    \item \brev{The autoencoder is trained with 20,000 samples from the Oxford dataset, distinct (in terms of traffic, weather, and locations) from those used for training the rest of \sysname.}
    \item The angles of the rotated anchors used by the RPN are set to -90$^{\circ}$, -45$^{\circ}$, 0$^{\circ}$, and 45$^{\circ}$.
    \item Both lidar and radar feature extractors are composed of four consecutive convolutional layers with a kernel size of 3 and padding of 1.
    \item The aspect ratio of the anchors is set to 2.5 to conform to the length-width ratio of regular vehicles~\cite{corolla}.
    \item The IoU threshold \newrev{(defined later)} of NMS for removing excessive proposals during testing is set to 0.2.
\end{itemize}
In the local training process, we employ the SGD optimizer by setting the initial learning rate as 0.01 and the decay factor as 0.01.

\begin{figure*}[t]
    \setlength\abovecaptionskip{6pt}
    \vspace{-1.5ex}
	   \captionsetup[subfigure]{justification=centering}
		\centering
		\subfloat[AP@IoU=0.5.]{
		    \begin{minipage}[b]{0.24\linewidth}
		        \centering
			    \includegraphics[width = 0.96\textwidth]{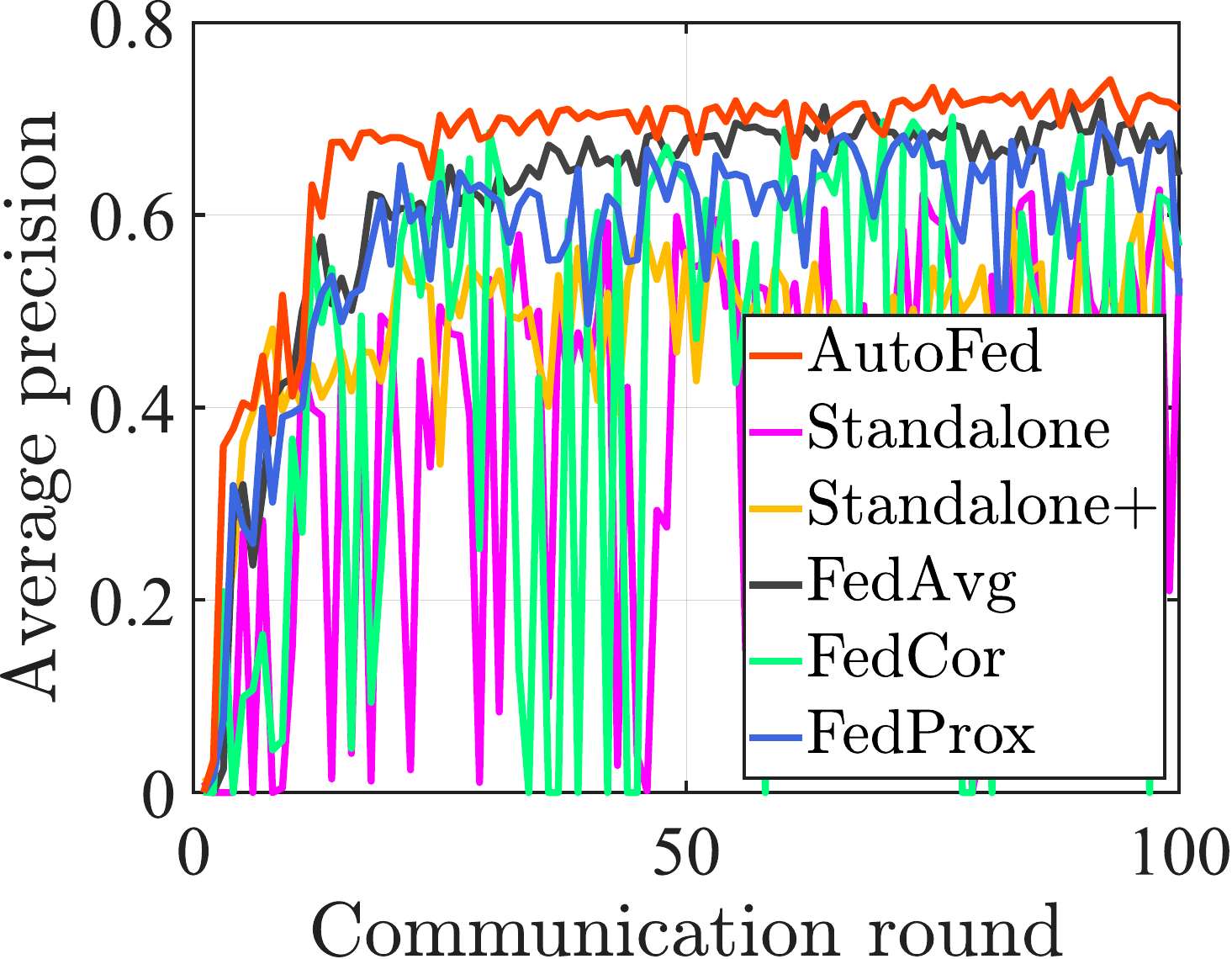}
			    \label{subfig:ap_iou05}
			\end{minipage}
		}
		\subfloat[AP@IoU=0.65.]{
		  \begin{minipage}[b]{0.24\linewidth}
		        \centering
			    \includegraphics[width = 0.96\textwidth]{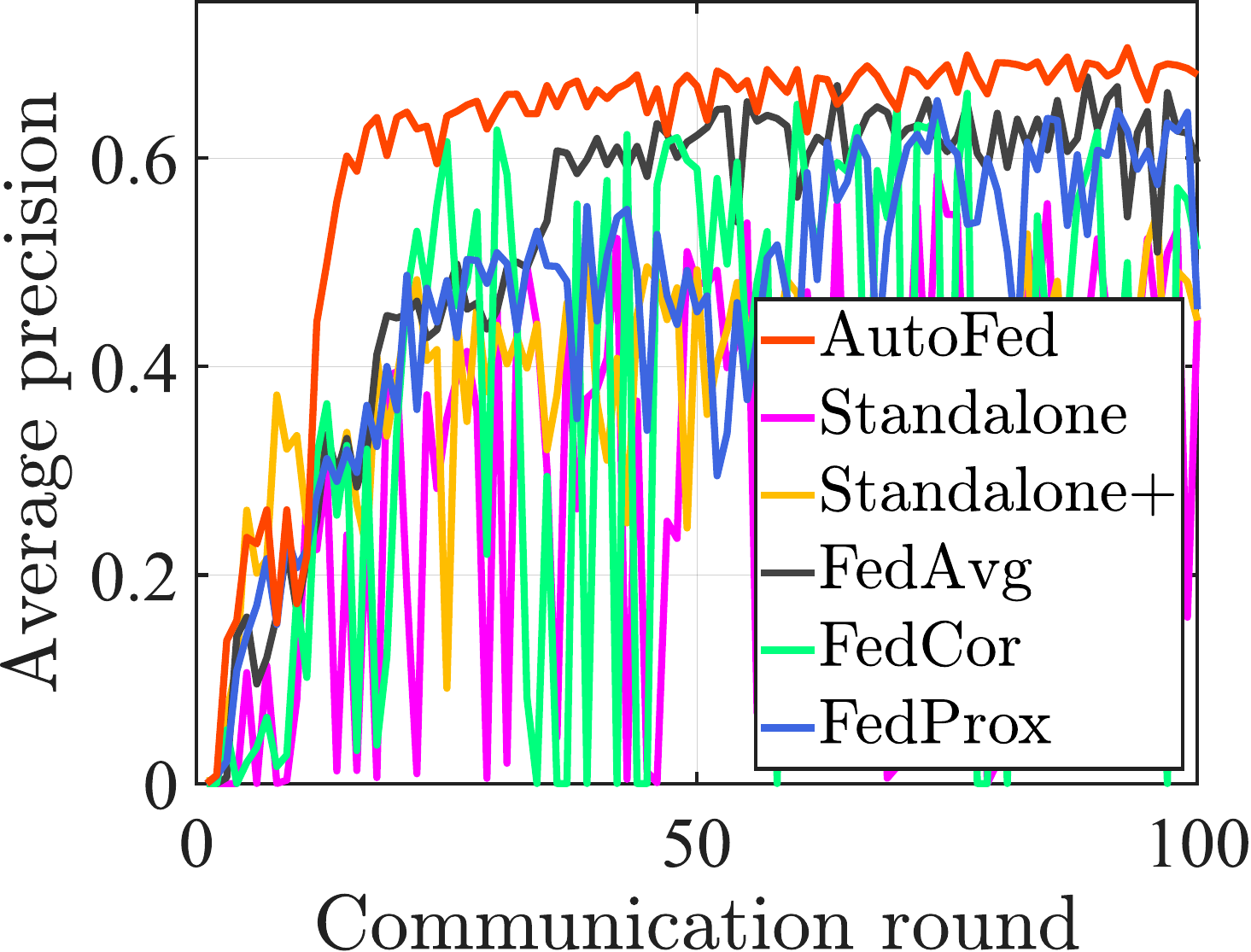}
			    \label{subfig:ap_iou065}
			\end{minipage}
		}
		\subfloat[AP@IoU=0.8.]{
		    \begin{minipage}[b]{0.24\linewidth}
		        \centering
			    \includegraphics[width = 0.96\textwidth]{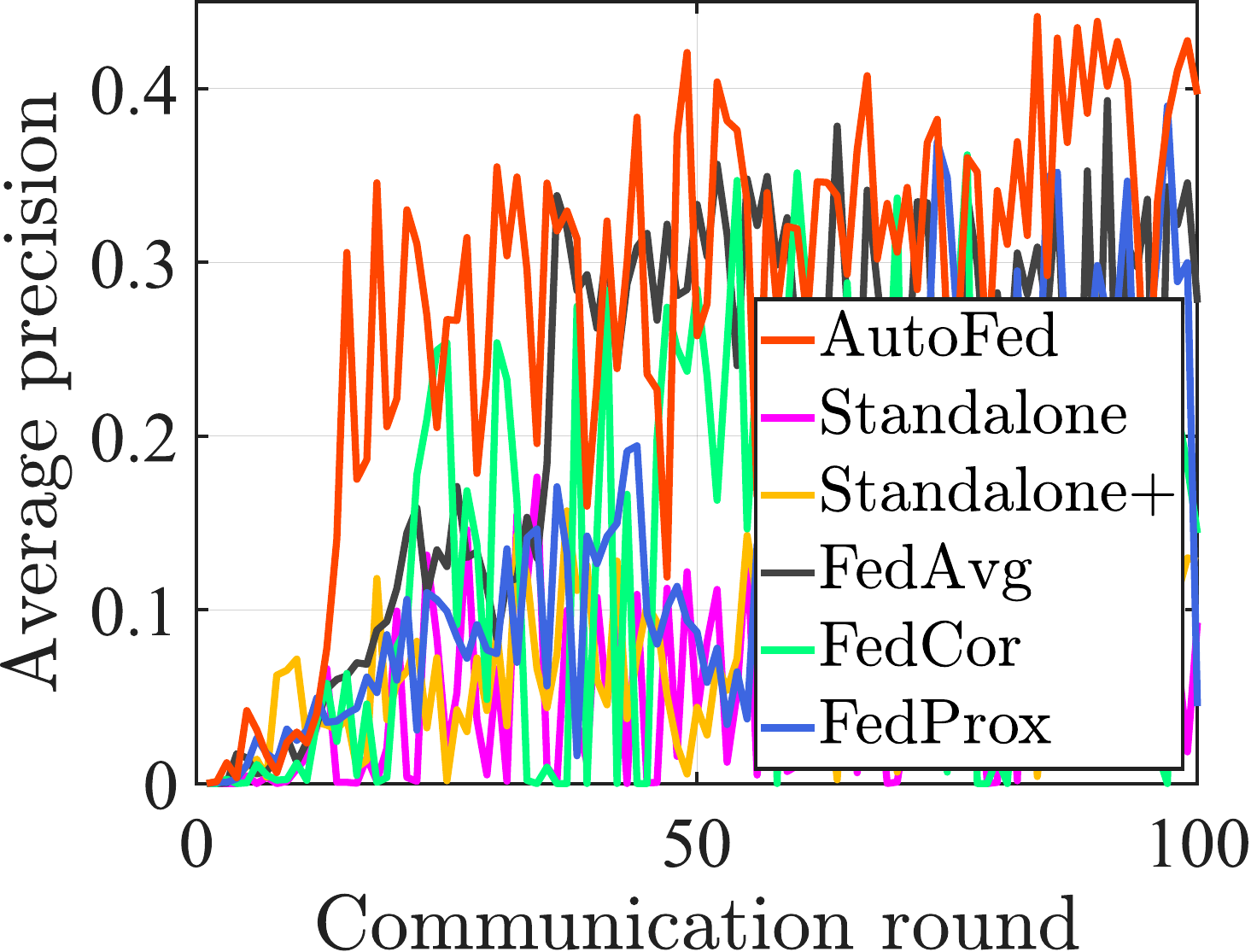}
			    \label{subfig:ap_iou08}
			\end{minipage}
		}
		\subfloat[AP@IoU=0.5:0.9.]{
		  \begin{minipage}[b]{0.24\linewidth}
		        \centering
			    \includegraphics[width = 0.96\textwidth]{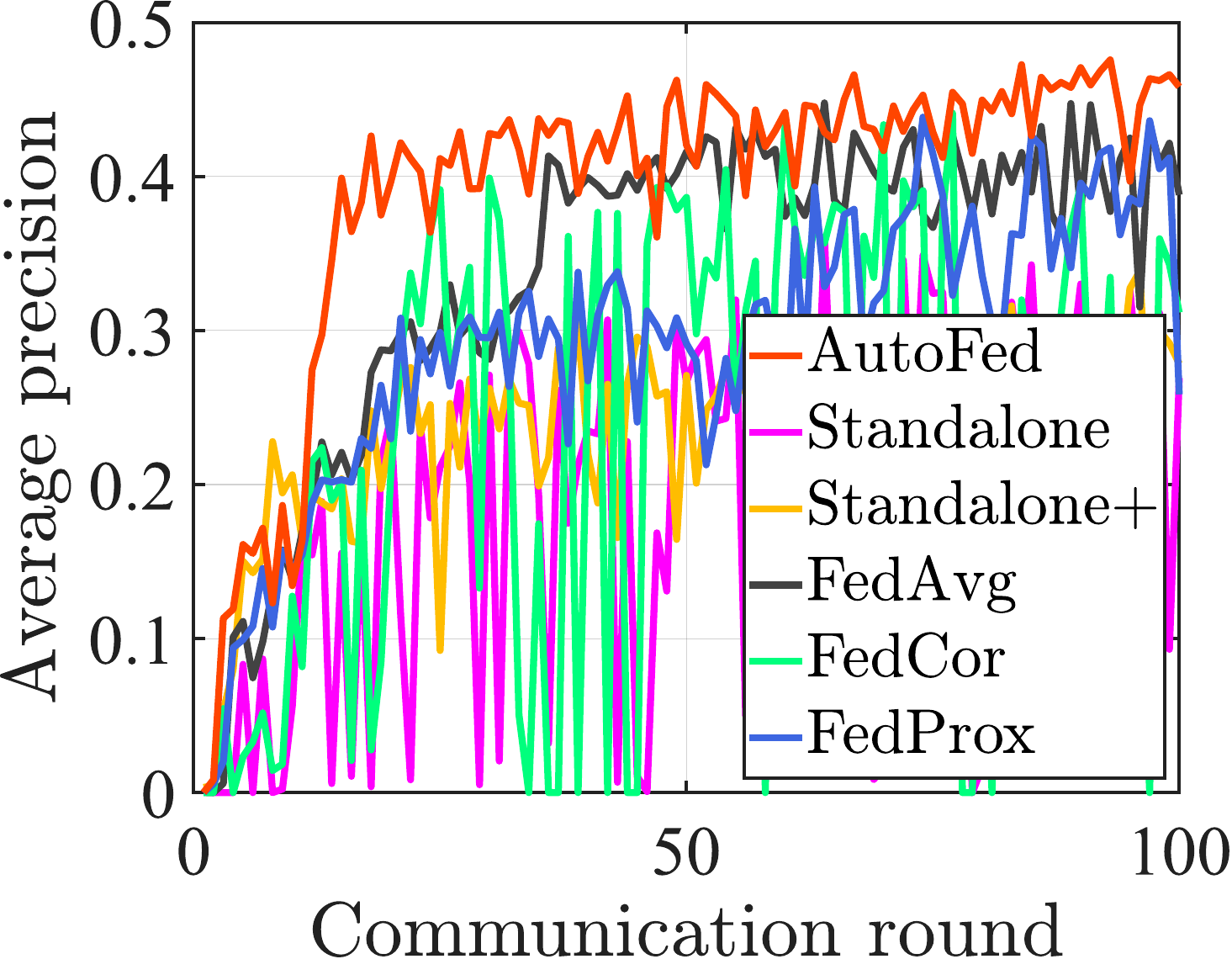}
			    \label{subfig:ap_iou0509}
			\end{minipage}
		}
		\\ \vspace{-1.6ex}
		\subfloat[AR, maxDets = 1.]{
		  \begin{minipage}[b]{0.24\linewidth}
		        \centering
			    \includegraphics[width = 0.96\textwidth]{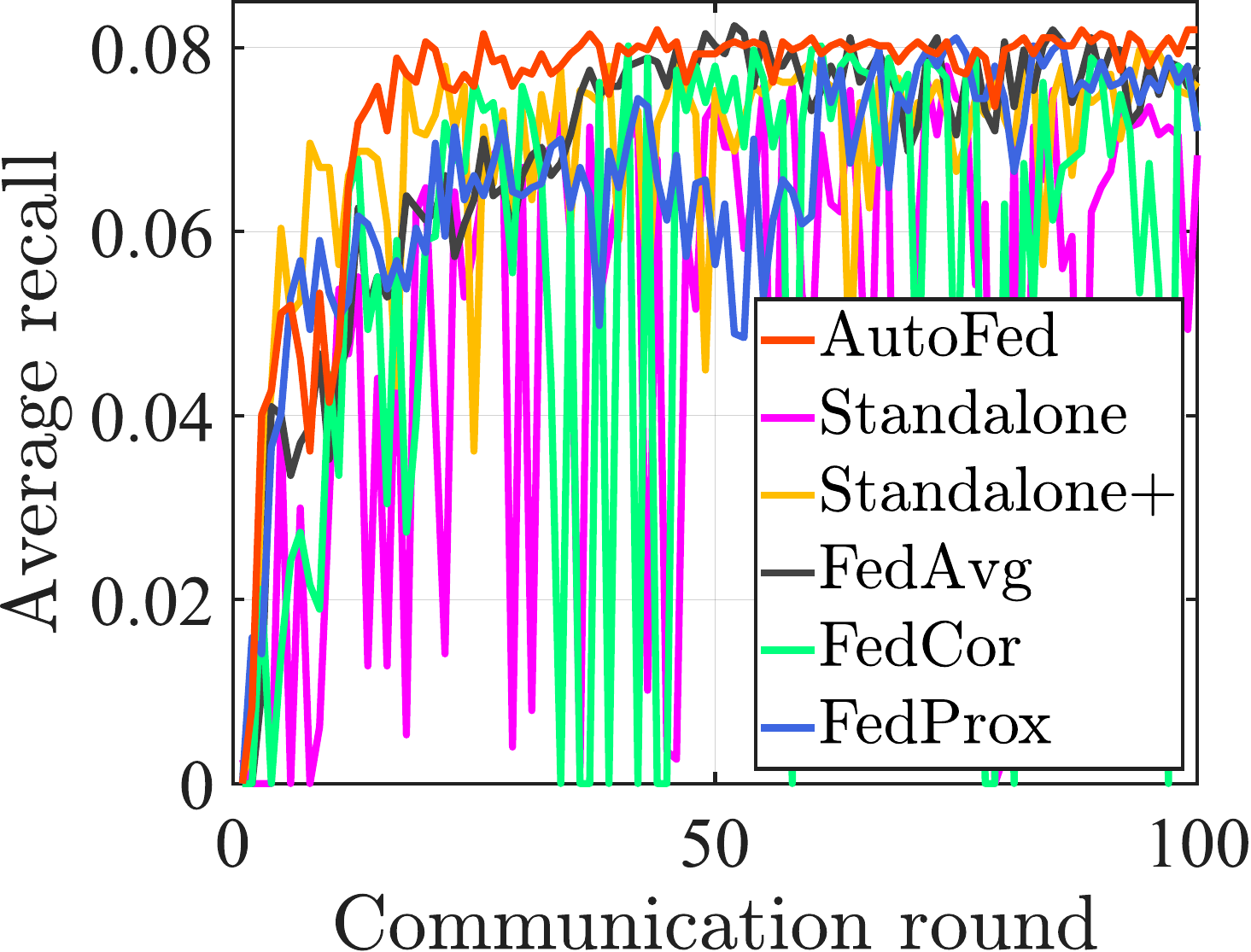}
			    \label{subfig:ar_max1}
			\end{minipage}
		}
		\subfloat[AR, maxDets = 10.]{
		    \begin{minipage}[b]{0.24\linewidth}
		        \centering
			    \includegraphics[width = 0.96\textwidth]{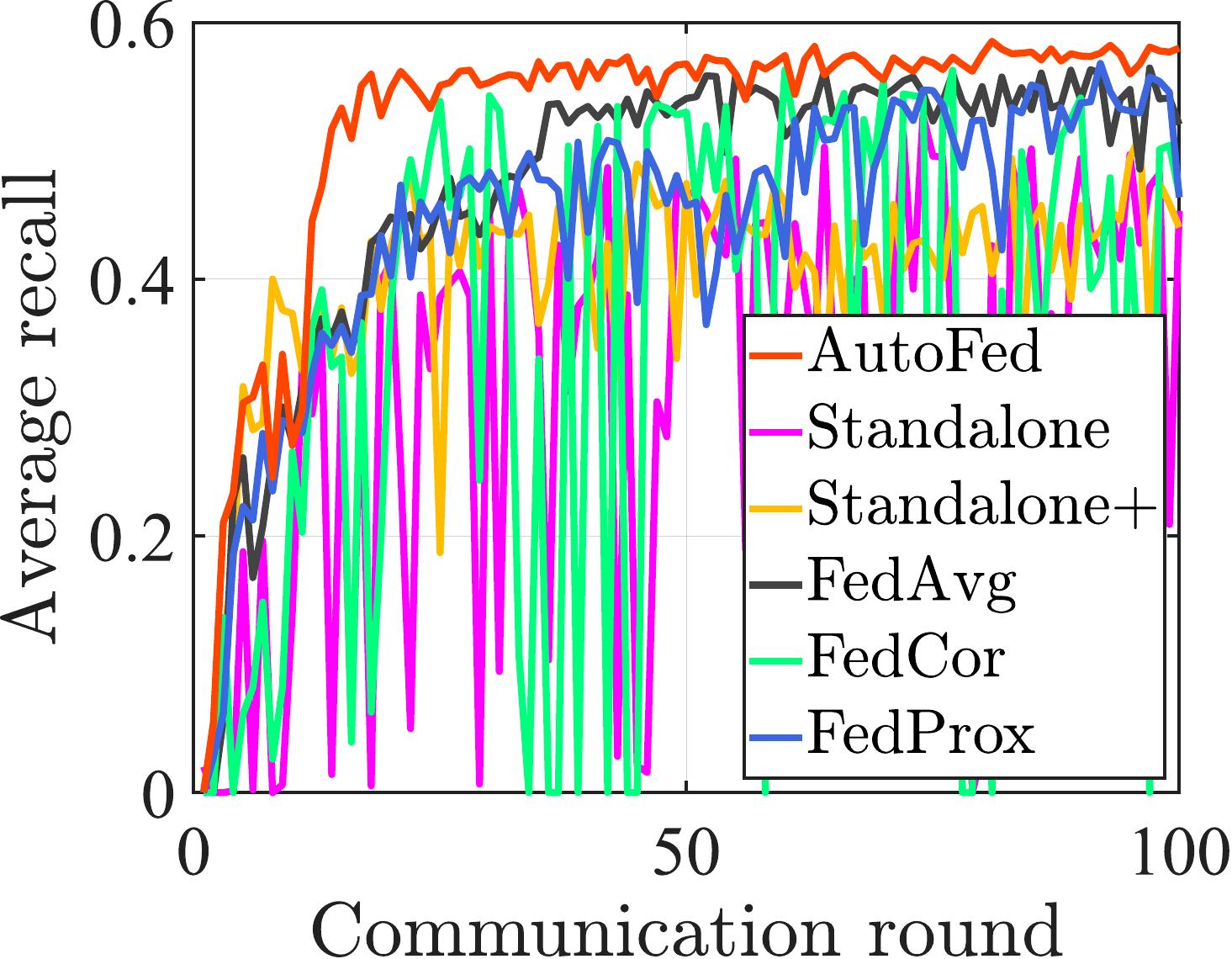}
			    \label{subfig:ar_max10}
			\end{minipage}
		}
		\subfloat[AR, maxDets = 100.]{
		  \begin{minipage}[b]{0.24\linewidth}
		        \centering
			    \includegraphics[width = 0.96\textwidth]{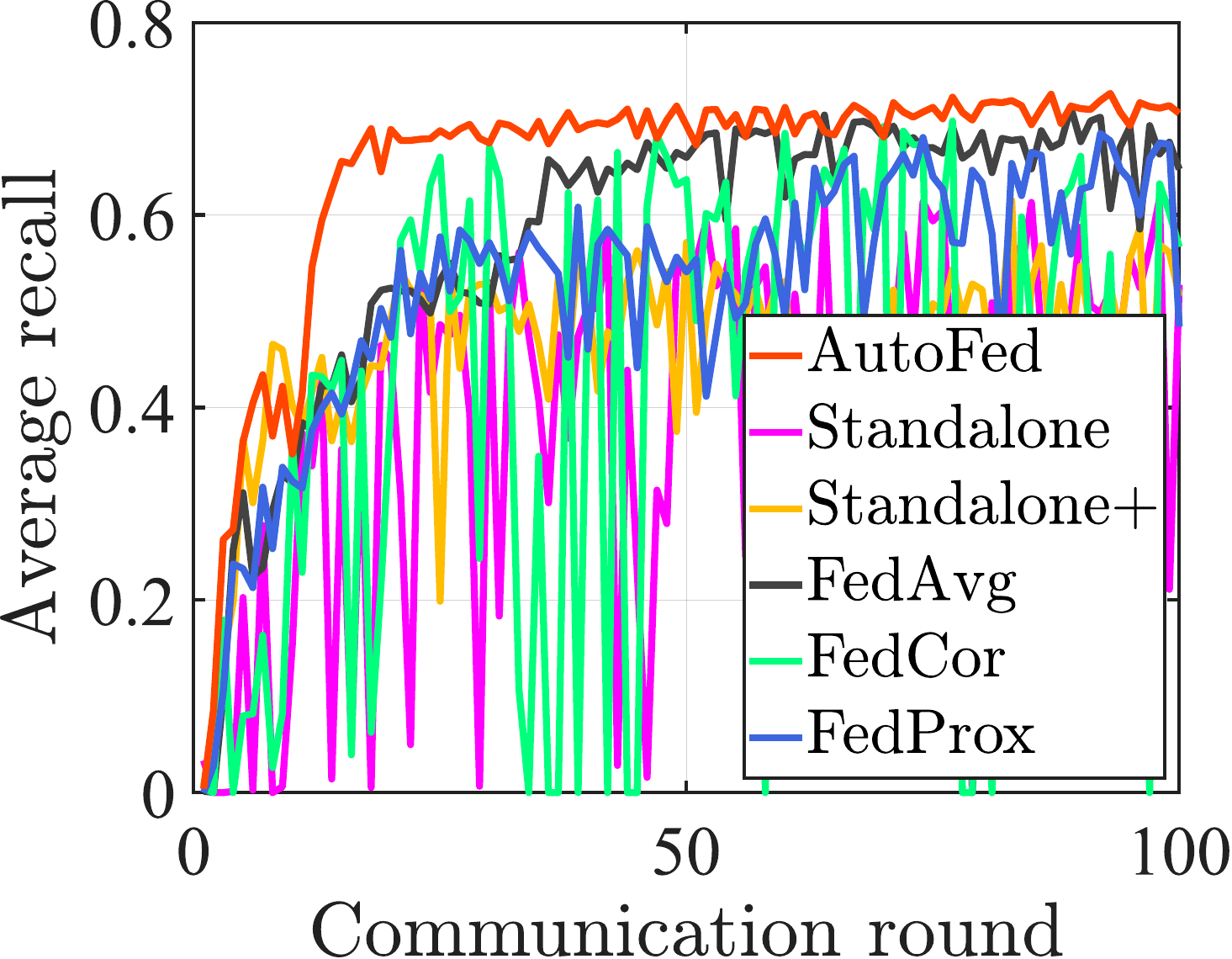}
			    \label{subfig:ar_max100}
			\end{minipage}
		}
		\subfloat[Convergence time.]{
		    \begin{minipage}[b]{0.24\linewidth}
		        \centering
			    \includegraphics[width = 0.96\textwidth]{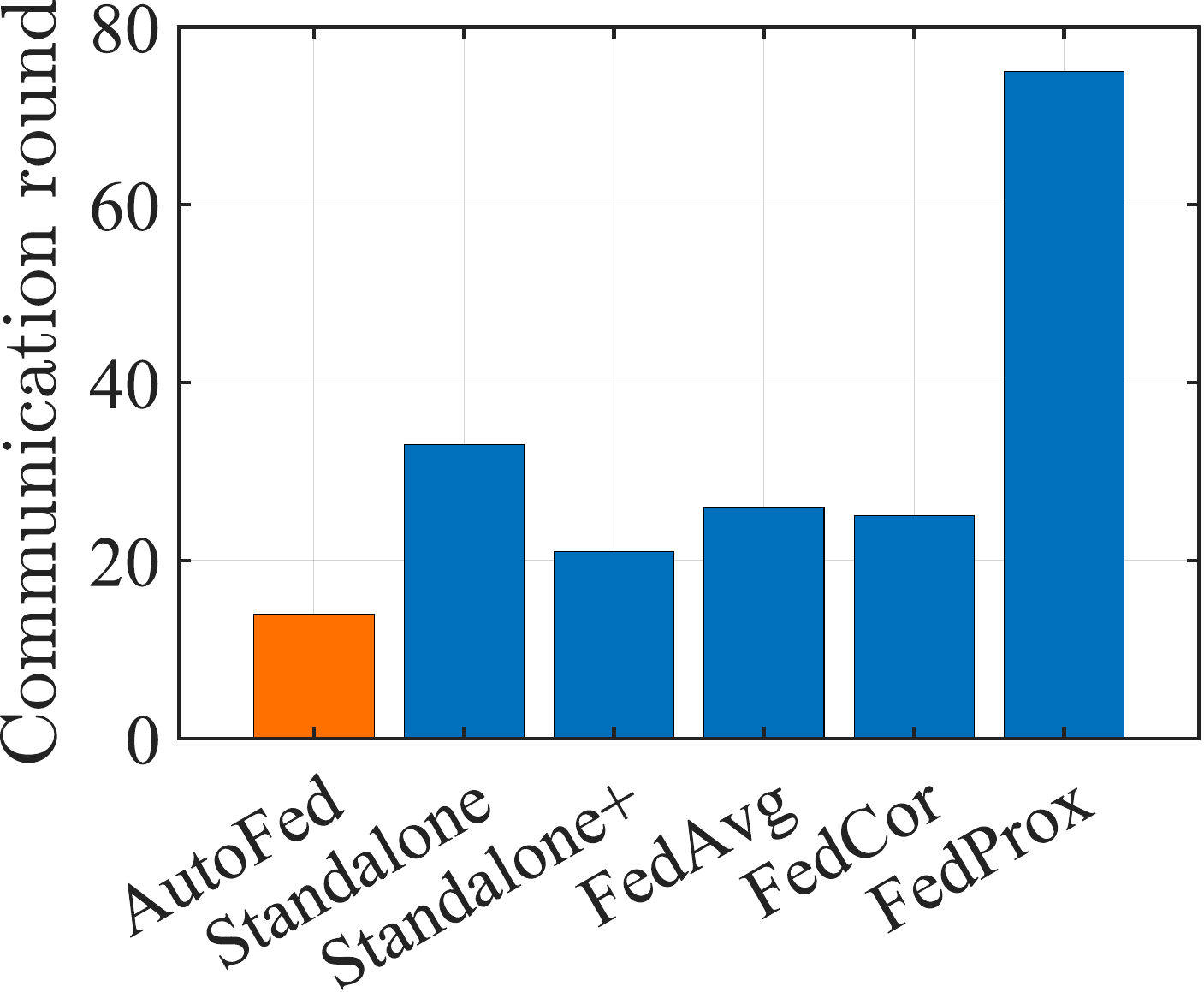}
			    \label{subfig:convergence}
			\end{minipage}
		}
		\caption{\brev{Comparing \sysname\ with several baseline methods, in terms of FL convergence and communication overhead.}}
		\label{fig:baseline_comparison}
\end{figure*}

\subsection{Experiment Setup} \label{ssec:setup}

\paragraph{Baselines} To comprehensively evaluate the performance of \sysname, we compare \sysname\ against five baselines:
\begin{itemize}
    \item \textbf{Standalone} trains a vehicle detection model using heterogeneous data (e.g., heterogeneous annotations, sensing modalities, and environments) locally without collaborations among clients. 
    \item \textbf{Standalone+} trains a vehicle detection model locally using the same setting as Standalone, but the data are sampled in a homogeneous way.
    \item \textbf{FedAvg} is the first and perhaps the most widely adopted FL method~\cite{mcmahan2017communication}. During training, all clients communicate updated local parameters to the central server and download the aggregated (i.e., averaged) global model for local training in the next round.
    \item \textbf{FedCor} is a correlation-based client selection strategy for heterogeneous FL~\cite{tang2022fedcor}. It formulates the goal of accelerating FL convergence as \newrev{an optimization problems that maximizes} the posterior expectation of loss decrease utilizing the Gaussian process.
    \item \textbf{FedProx} adds a proximal term to the loss function of local training to reduce the distance between the local model and the global model~\cite{li2020federated}, hence addressing both system and statistical heterogeneity.
\end{itemize}
In addition, we adopt the same multimodal vehicle detection model configuration for each baseline method as \sysname. We also apply the same training settings and data configurations as \sysname\ to the baseline methods, the results are reported after the same number of communication rounds. \brev{It should be noted that we use Standalone and Standalone+ as baselines to provide context for how better FL methods perform: it confirms that they do improve upon standalone training, because each client only has limited data in reality.}

\paragraph{Evaluation Metrics.} Before introducing the evaluation metrics, we first define an important concept called IoU (intersection over union), which evaluates the overlap between two bounding boxes. Suppose the ground truth and predicted bounding boxes are $B_{gt}$ and $B_p$, respectively, then IoU is given by the overlapping area between the predicted bounding box and the ground truth bounding box divided by the area of union between them:
\begin{align}
    \mathrm{IoU} = \frac{\mathrm{Area}(B_{p}\cap B_{gt})}{\mathrm{Area}(B_{p}\cup B_{gt})}
\end{align}
We define $\mathit{TP}$ as the number of correct detections (i.e., detections with an IoU greater than the predefined threshold), $\mathit{FP}$ as wrong detections (i.e.,  detections with an IoU smaller than the threshold), and $\mathit{FN}$ as the number of ground truths that are not identified. Based on these definitions, we define precision and recall as:
\begin{align}
    \mathrm{Precision} = \frac{\mathit{TP}}{\mathit{TP}+\mathit{FP}},\quad \mathrm{Recall} = \frac{\mathit{TP}}{\mathit{TP}+\mathit{FN}}.
\end{align}
Since there is often a tradeoff between precision and recall, we also define an average precision (AP) value across all precision values from 0 to 1, thus summarizing the precision-recall curve. Moreover, we calculate the average recall (AR) value at IoU thresholds from 0.5 to 1, thus summarizing the distribution of recall values across a range of IoU thresholds~\cite{lin2014microsoft}. AP and AR are our key evaluation metrics hereafter.

\subsection{Superiority of \sysname} \label{ssec:superiority}

\begin{figure*}[t]
    \setlength\abovecaptionskip{6pt}
    \vspace{-2.5ex}
	   \captionsetup[subfigure]{justification=centering}
		\centering
		\subfloat[Ground truth.]{
		  \begin{minipage}[b]{0.135\linewidth}
		        \centering
			    \includegraphics[width = 0.99\textwidth]{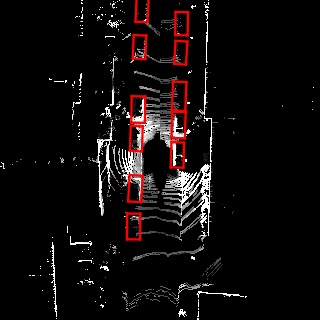}
			    \label{subfig:de_gt}
			\end{minipage}
		}
		\subfloat[\sysname.]{
		  \begin{minipage}[b]{0.135\linewidth}
		        \centering
			    \includegraphics[width = 0.99\textwidth]{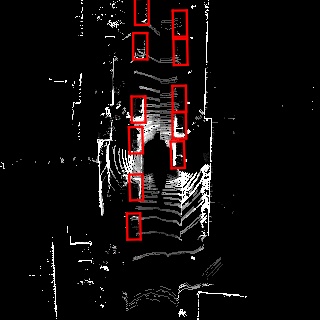}
			    \label{subfig:de_fedauto}
			\end{minipage}
		}
  	\subfloat[\brev{Standalone.}]{
		  \begin{minipage}[b]{0.135\linewidth}
		        \centering
			    \includegraphics[width = 0.99\textwidth]{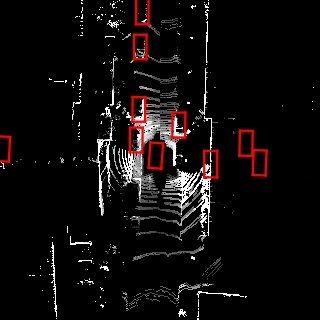}
			    \label{subfig:de_mono}
			\end{minipage}
		}
		\subfloat[Standalone+.]{
		  \begin{minipage}[b]{0.135\linewidth}
		        \centering
			    \includegraphics[width = 0.99\textwidth]{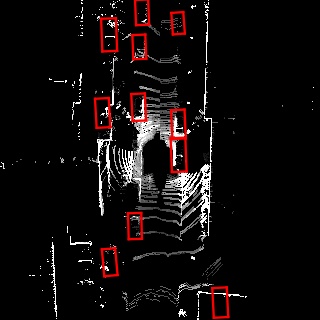}
			    \label{subfig:de_mono_plus}
			\end{minipage}
		}
		\subfloat[FedAvg.]{
		    \begin{minipage}[b]{0.135\linewidth}
		        \centering
			    \includegraphics[width = 0.99\textwidth]{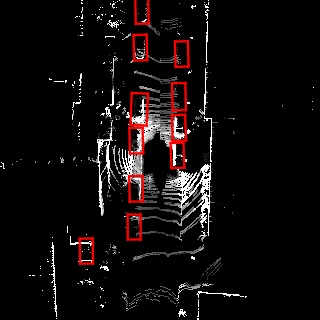}
			    \label{subfig:de_fedavg}
			\end{minipage}
		}
		\subfloat[FedCor.]{
		  \begin{minipage}[b]{0.135\linewidth}
		        \centering
			    \includegraphics[width = 0.99\textwidth]{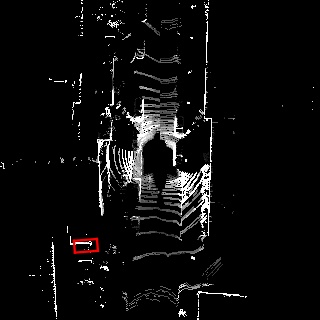}
			    \label{subfig:de_fedcor}
			\end{minipage}
		}
		\subfloat[FedProx.]{
		    \begin{minipage}[b]{0.135\linewidth}
		        \centering
			    \includegraphics[width = 0.99\textwidth]{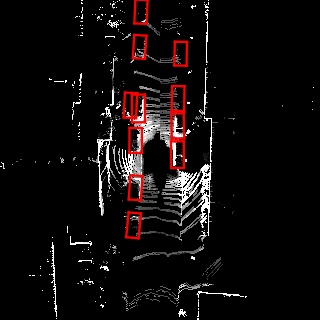}
			    \label{subfig:de_FedProx}
			\end{minipage}
		}
		\caption{Example detection results of \sysname\ and other baseline methods.}
		\label{fig:detection_examples}
	    \vspace{-.5ex}
\end{figure*}

We compare \sysname\ with the baselines in terms of the evaluation metrics defined in \S~\ref{ssec:setup}. Specifically, we report AP when the IoU is 0.5, 0.65, and 0.8, respectively, and the mean AP when the IoU ranges from 0.5 to 0.9. As for AR, we focus on the cases when the number of maximum detections is 1, 10, and 100, respectively. We report the evaluation results in Figure~\ref{fig:baseline_comparison}. Figure~\ref{subfig:ap_iou05} shows that, when the IoU is set as 0.5, the AP of \sysname\ is 0.71 while the number of FedAvg and FedProx are 0.68 and 0.58, respectively. Moreover, the APs of Standalone, Standalone+, and FedCor oscillate dramatically  and barely converge. Similarly, as shown in Figures~\ref{subfig:ap_iou065},\brev{~\ref{subfig:ap_iou08}},  and~\ref{subfig:ap_iou0509}, the performance of \sysname\ significantly outperforms the baselines. \newrev{It might be curious that the AP curve of \sysname\ in Figure~\ref{subfig:ap_iou08} appears to be fluctuating, but this can be readily attributed to the fact} that setting IoU as 0.8 is a stringent criterion for the vehicle detection task and causes the performance to become unstable. 

With regard to AR \newrev{shown by} Figures ~\ref{subfig:ar_max1},~\ref{subfig:ar_max10}, and~\ref{subfig:ar_max100}, AutoFed \newrev{exhibits significantly better performance compared with the baselines, in terms of}
both AP and AR. Moreover, we also find \newrev{that, when compared with the baselines, \sysname\ reaches the maximum AP and AR with less number of communication rounds, as also confirmed by the results} presented in Figure~\ref{fig:baseline_comparison}. Specifically, while \sysname\ converges in 10 communication rounds, 
\newrev{all baseline} methods converge after 20 communication rounds. Furthermore, the AP and AR curves of \sysname\ rarely fluctuate, and the 
training of \sysname\ is much more stable than the baselines, indicating that the multimodal network trained by \sysname\ is much more robust.

We also \newrev{showcase} some examples of vehicle detection  in Figure~\ref{fig:detection_examples}. In the examples, we use the 2-D lidar intensity map as background for reference, and draw the ground truth and predicted bounding boxes upon it. Figure~\ref{subfig:de_fedauto} shows that \sysname\ generates high-precision vehicle detection results \newrev{very close} to the ground truth in Figure~\ref{subfig:de_gt}. In contrast, the \brev{Standalone}, Standalone+, and FedAvg methods make incorrect predictions outside the road, FedCor's  misses most of the vehicles, and FedProx misses some vehicles and generates inaccurate bounding boxes overlapped with each other. The results \newrev{evidently confirm} that \sysname\ outperforms the baselines with more accurate predictions.

Furthermore, we compare the communication cost of \newrev{\sysname\ training (the same as other FL baselines)} with centralized training, i.e., all the clients transfer the collected data  to a central server for training the model. The results show that, while centralized training transfers 660000~\!KB of sensor data during each communication round per client, \sysname\ only transfers 62246~\!KB of model weights. \newrev{In other words,} \sysname\
reduces up to more than 10$\times$ communication cost per client than the centralized training, firmly validating its communication-efficient design. 

\begin{figure}[h]
    \setlength\abovecaptionskip{6pt}
    \vspace{-2.5ex}
	   \captionsetup[subfigure]{justification=centering}
		\centering
		\subfloat[Average precision.]{
		  \begin{minipage}[b]{0.49\linewidth}
		        \centering
			    \includegraphics[width = 0.96\textwidth]{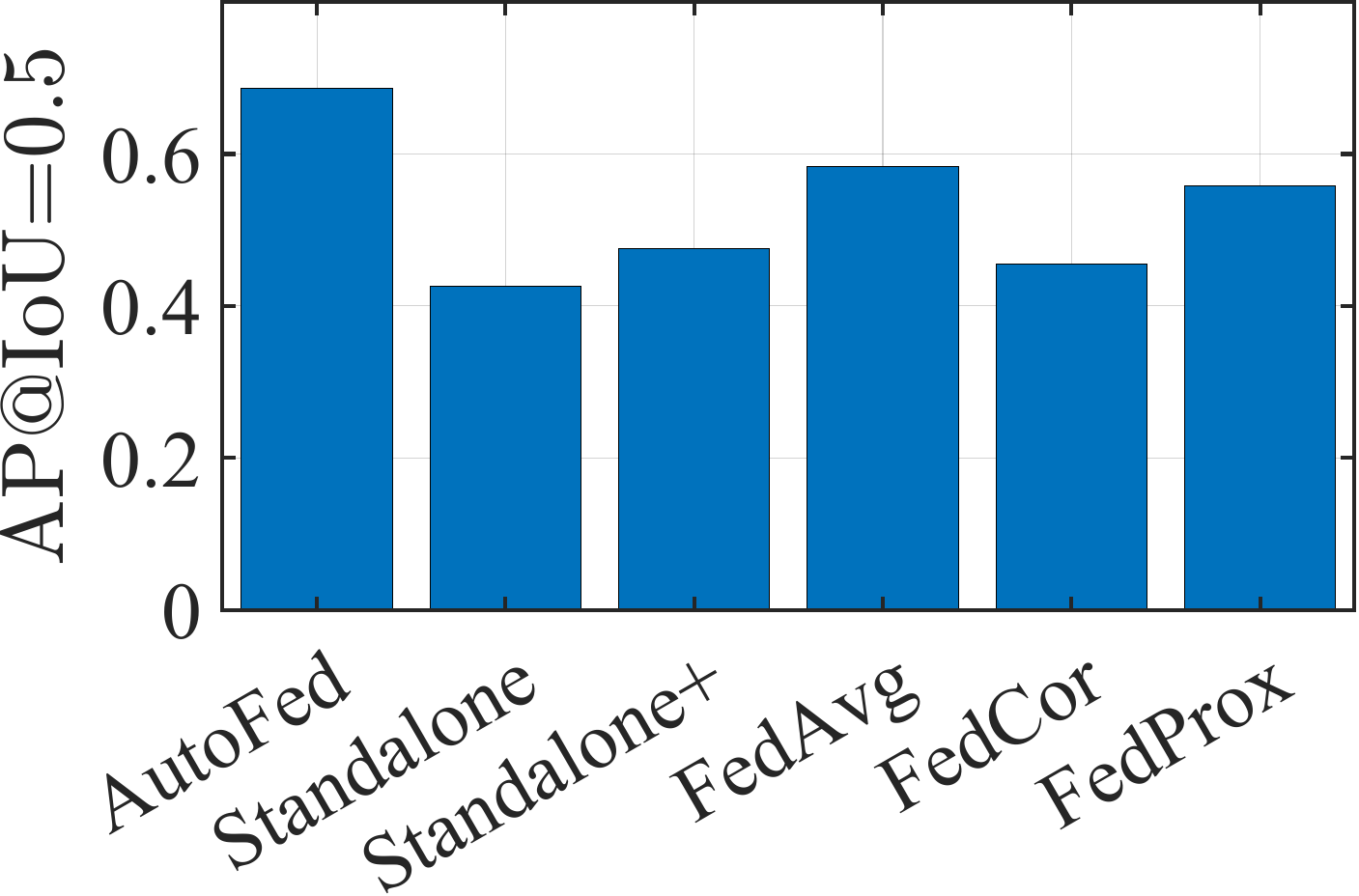}
			\end{minipage}
		    \label{subfig:nuscenes_ap}
		}
		\subfloat[Average recall.]{
		    \begin{minipage}[b]{0.49\linewidth}
		        \centering
			    \includegraphics[width = 0.96\textwidth]{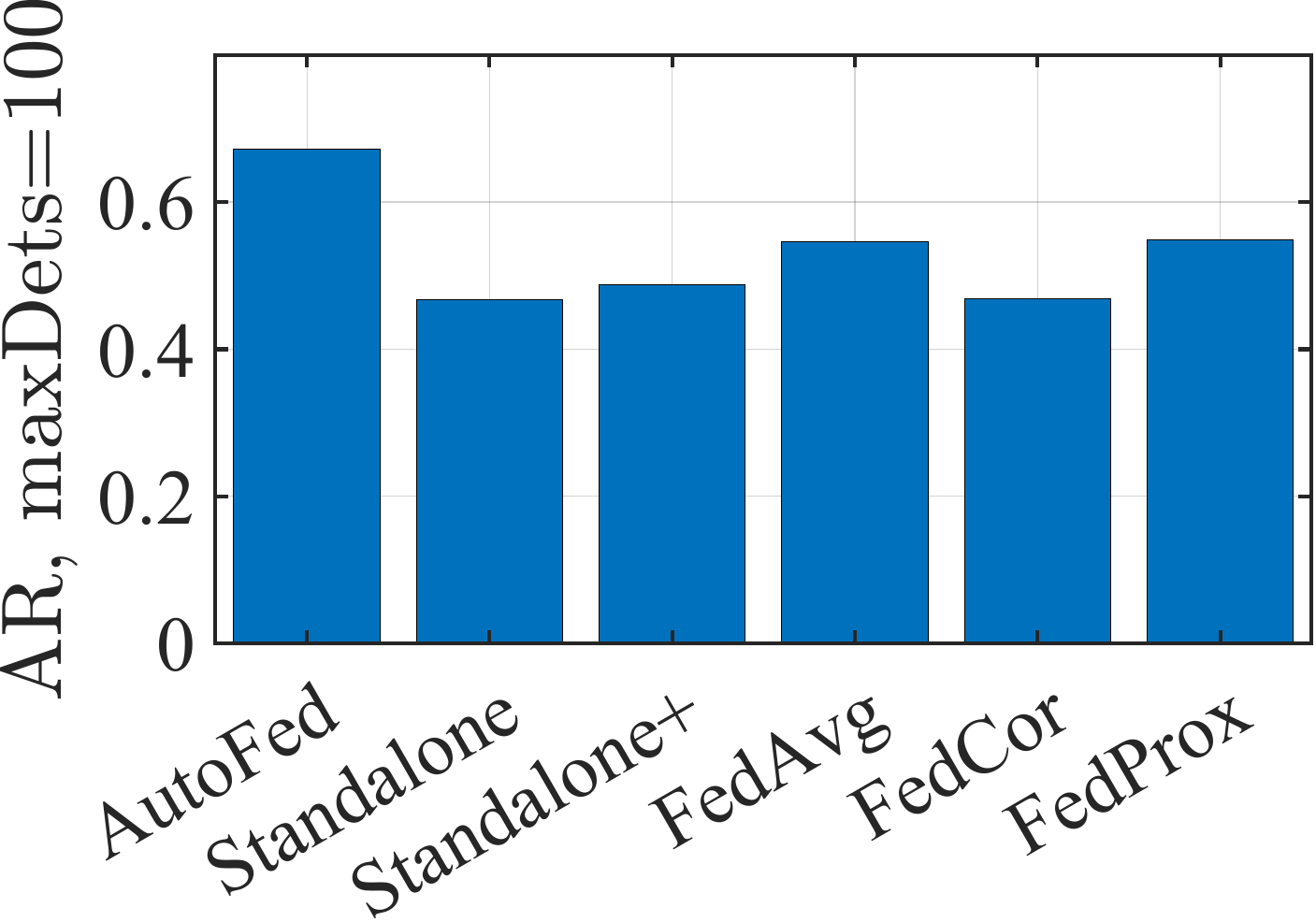}
			\end{minipage}
			\label{subfig:nuscenes_ar}
		}
		\caption{\brev{Evaluation on the nuScenes dataset.}}
		\label{fig:eval_nuscenes}
		\vspace{-1.5ex}
\end{figure}

\brev{We finally compare the performance of \sysname\ with the baselines on the nuScenes dataset~\cite{caesar2020nuscenes} to demonstrate its generalizability across different datasets. We train \sysname\ for 100 communication rounds on the dataset. As shown in Figure~\ref{fig:eval_nuscenes}, \sysname\ outperforms all of the baselines on the nuScenes dataset by a large margin, firmly demonstrating that the evaluation results can be generalized to other datasets as well. It is worth noting that the overall AP and AR results of \sysname\ on this dataset (0.687 and 0.672) are slightly lower than those shown in Figures~\ref{subfig:ap_iou05} and \ref{subfig:ar_max100} on the Oxford Radar RobotCar dataset, which can be attributed to a variety of factors, such as the complexity of the scenes and objects, sensor mounting positions, and most importantly, the sparsity and lower quality of the radar point cloud provided by the nuScenes dataset. }

\subsection{Cross-domain Robustness}
We evaluate the robustness of \sysname in cross-domain settings by investigating how the trained model performs in \newrev{varied sensing modalities and different weather conditions.} Since the AVs' routes in the experiment encompass different roads and areas, the results in \S~\ref{ssec:superiority} have already proven the cross-road and cross-area capabilities of \sysname, therefore we omit their discussions here. %

\subsubsection{Various Sensing Modalities}
\begin{figure}[b]
    \setlength\abovecaptionskip{6pt}
    \vspace{-2.5ex}
	   \captionsetup[subfigure]{justification=centering}
		\centering
		\subfloat[Average precision.]{
		  \begin{minipage}[b]{\linewidth}
		        \centering
			    \includegraphics[width = 0.96\textwidth]{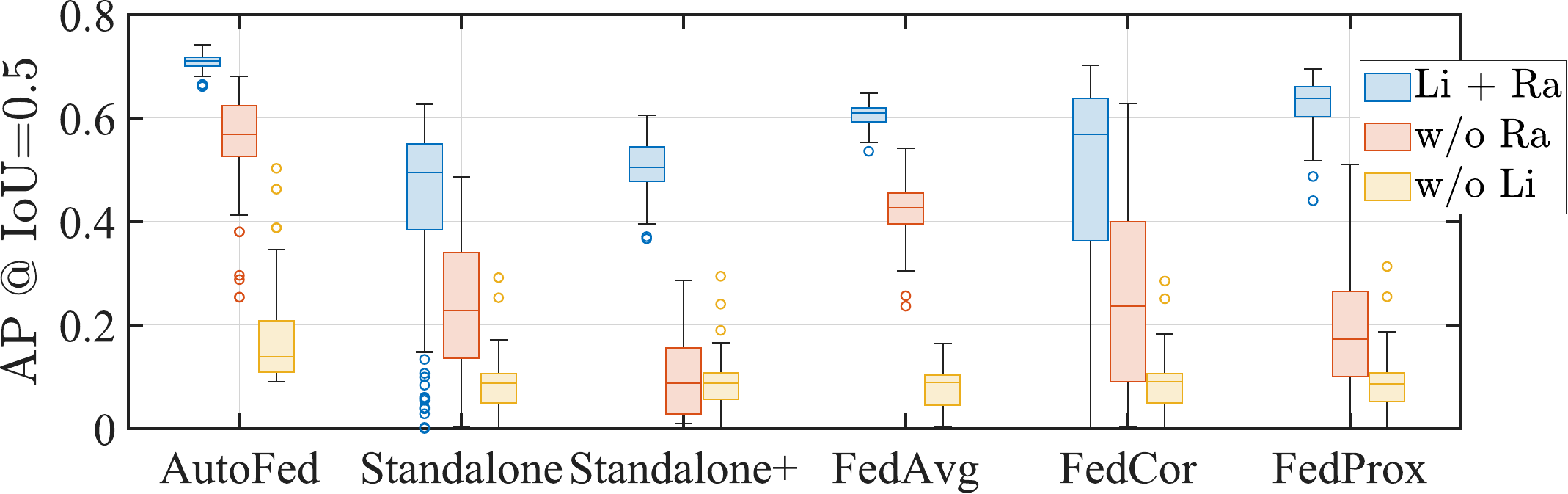}
			    \label{subfig:sensor_precision}
			\end{minipage}
		}
		\\
		\subfloat[Average recall.]{
		  \begin{minipage}[b]{\linewidth}
		        \centering
			    \includegraphics[width = 0.96\textwidth]{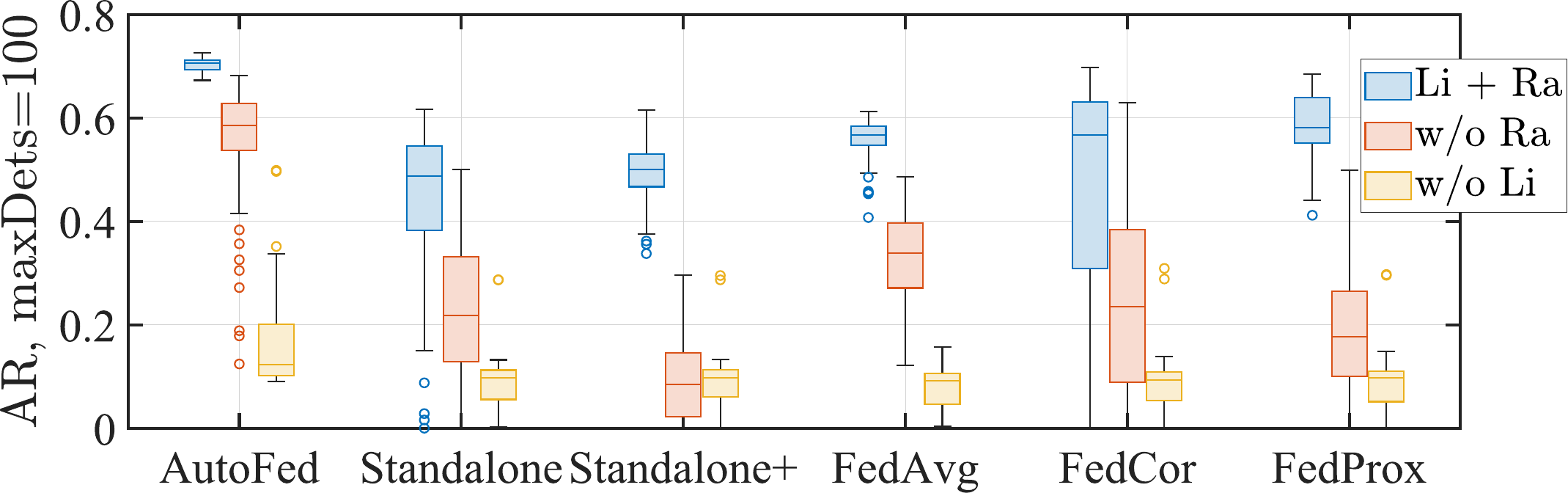}
			    \label{subfig:sensor_recall}
			\end{minipage}
		}
		\\
		\subfloat[Lidar + radar.]{
		    \begin{minipage}[b]{0.33\linewidth}
		        \centering
			    \includegraphics[width = 0.96\textwidth]{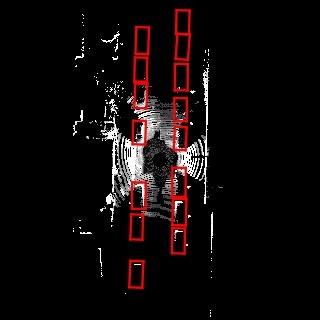}
			    \label{subfig:li_and_ra}
			\end{minipage}
		}
		\subfloat[Missing radar.]{
		    \begin{minipage}[b]{0.33\linewidth}
		        \centering
			    \includegraphics[width = 0.96\textwidth]{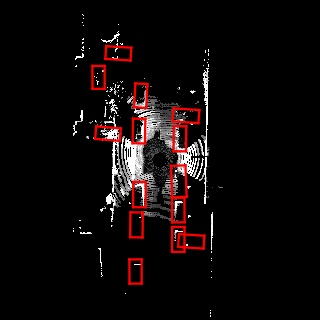}
			    \label{subfig:wo_ra}
			\end{minipage}
		}
		\subfloat[Missing lidar.]{
		    \begin{minipage}[b]{0.33\linewidth}
		        \centering
			    \includegraphics[width = 0.96\textwidth]{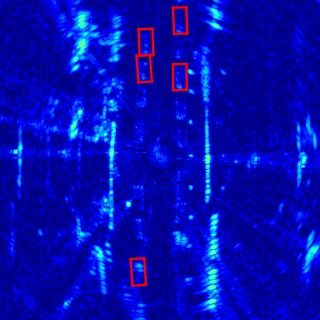}
			    \label{subfig:wo_li}
			\end{minipage}
		}
		\caption{Different missing modalities.}
		\label{fig:eval_sensing_modality}
		\vspace{-1.5ex}
\end{figure}

Since \sysname\ involves both lidar and radar sensors, there are three possible sensor combinations, i.e., i) lidar + radar (Li + Ra), ii) without radar (w/o Ra), and iii) without lidar (w/o Li). We evaluate the performance of \sysname\  under these three settings, and report the results in Figure~\ref{fig:eval_sensing_modality}. The results show that, when the IoU is set to be above 0.5, the median APs achieved by \sysname\ are 0.71, 0.57, and 0.12 under the aforementioned three settings. Correspondingly, the median ARs achieved by \sysname\ are 0.70, 0.59, and 0.12.
The autoencoder employed by \sysname\ helps the model to maximize the efficacy of information embedded in either radar or lidar data, and \sysname\ \newrev{exhibits} the smallest performance drop compared with the  baselines whose performance is drastically impacted by missing modalities. However, since the performance drop of missing modalities stems from the loss of information, even the adoption of an autoencoder cannot \newrev{totally} fill up the performance gap. \brev{We have also noticed that the AP and AR of \sysname\ are significantly lower in the radar-only mode compared to the other sensor combinations. Upon further investigation, we suspect that this may be because the importance of radar is overshadowed by lidar that provides most of the information used by \sysname. Specifically, the majority of the vehicles in the dataset are close to the ego vehicle, probably due to the narrow width of the road, and as a result, lidar can detect almost all of these vehicles because they are within its range. This leads to the lower performance of the radar-only mode, as radar is 
often meant to supplement the lidar sensor for long-range detection.}

We also show one example of vehicle detection with three sensor combinations in Figure~\ref{fig:eval_sensing_modality}. As Figure~\ref{subfig:li_and_ra} illustrates, when both lidar and radar are available, \sysname\ is able to recognize most of the vehicles on the road. As a comparison,  Figure~\ref{subfig:wo_ra} shows that  missing radar data affects the detection of vehicles in the \newrev{further} distance, but the nearby vehicles can still be identified. This phenomenon is consistent with the characteristics of the radar sensor, i.e., the radar has an extended range due to better penetration capability while lidar can only obtain a much shorter range due to attenuation caused by in-air particles~\cite{zhao2020method}. In addition, we also visualize the case of missing lidar in Figure~\ref{subfig:wo_li}, where the vehicles \newrev{in distance} can be well detected by the radar. The results \newrev{clearly demonstrate the complementary sensing capability of radar and lidar.}

\begin{figure}[b]
    \setlength\abovecaptionskip{6pt}
    \vspace{-2.5ex}
	   \captionsetup[subfigure]{justification=centering}
		\centering
		\subfloat[Average precision.]{
		  \begin{minipage}[b]{0.98\linewidth}
		        \centering
			    \includegraphics[width = 0.96\textwidth]{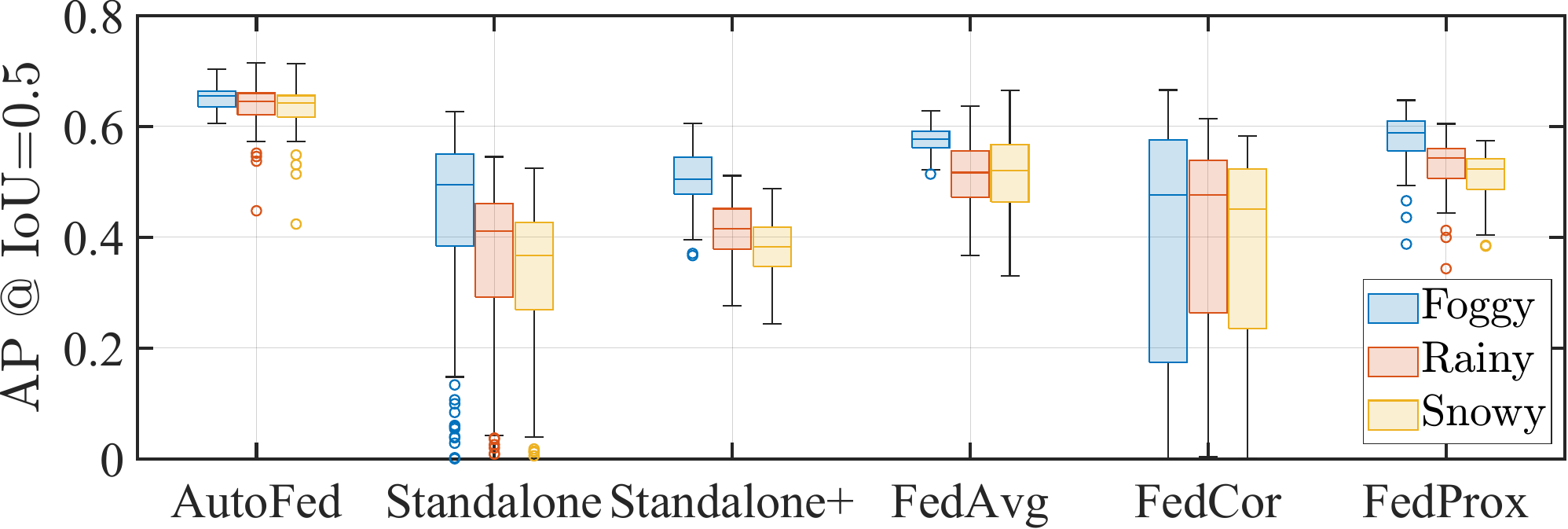}
			    \label{subfig:weather_precision}
			\end{minipage}
			
		}
		\\
		\subfloat[Average recall.]{
		  \begin{minipage}[b]{0.98\linewidth}
		        \centering
			    \includegraphics[width = 0.96\textwidth]{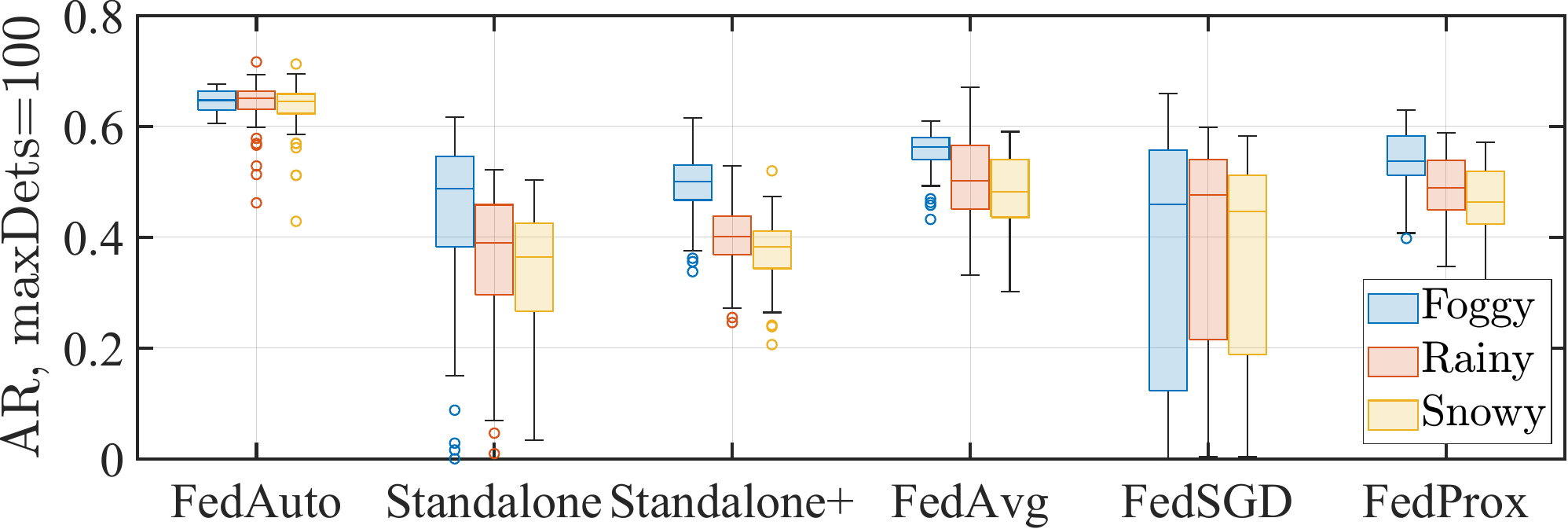}
			    \label{subfig:weather_recall}
			\end{minipage}
		}
		\\
		\subfloat[Foggy.]{
		    \begin{minipage}[b]{0.33\linewidth}
		        \centering
			    \includegraphics[width = 0.96\textwidth]{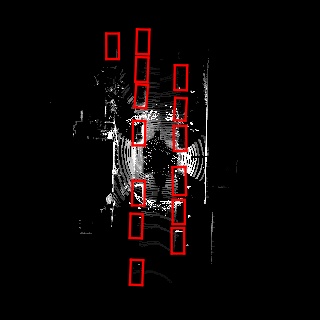}
			    \label{subfig:de_foggy}
			\end{minipage}
		}
		\subfloat[Rainy.]{
		    \begin{minipage}[b]{0.33\linewidth}
		        \centering
			    \includegraphics[width = 0.96\textwidth]{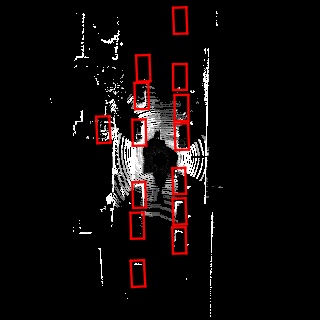}
			    \label{subfig:de_rainy}
			\end{minipage}
		}
		\subfloat[Snowy.]{
		    \begin{minipage}[b]{0.33\linewidth}
		        \centering
			    \includegraphics[width = 0.96\textwidth]{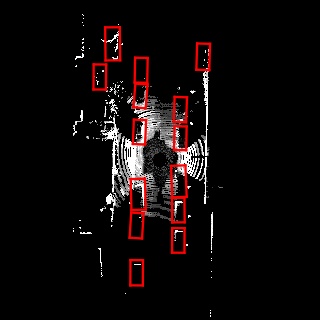}
			\end{minipage}
			\label{subfig:de_snowy}
		}
		\caption{Different weathers.}
		\label{fig:eval_weather}
		\vspace{-1.5ex}
\end{figure}

\subsubsection{Robustness against Adverse Weather Conditions}
Adverse weather is a realistic but challenging \newrev{scenario} for vehicle detection, which has a negative impact on the sensing capabilities~\cite{kilic2021lidar}. 
\newrev{Therefore,} we evaluate the performance of \sysname\ under different adverse weathers (e.g., foggy, rainy, and snowy). Due to the lack of available datasets collected under adverse weather, we employ the physical models in DEF~\cite{bijelic2020seeing} and LISA~\cite{kilic2021lidar} to simulate fog, rain, and snow respectively. Specifically, we set the fog density to 0.05m$^{-1}$ in the DEF model and the rate of rain and snow to 30~\!mm/h in the LISA model.
Comparing the backgrounds in Figures~\ref{subfig:de_foggy},~\ref{subfig:de_snowy}, and~\ref{subfig:de_rainy},  while foggy weather attenuates lidar signals and shrinks the field of view, rainy and snowy weathers mainly affect the lidar signals by inducing scattered reflections near the sensor. In particular, the three adverse weather conditions degrade the median AP of \sysname\ from 0.71 to 0.65, 0.63, and 0.63, respectively, and degrade the median AR from 0.71 to 0.64, 0.63, and 0.63, respectively. The performance discrepancies among these adverse weathers can be attributed to their different reflectance of lidar signals. Despite the performance degradation, \sysname\ exhibits the best generalization when compared with the baselines. \brev{The consistently high performance of \sysname\ under all adverse weather conditions confirms that the client selection mechanism has allowed the DNN model to effectively incorporate information from unusual circumstances after sufficient training.}

\subsection{Ablation Study}
We evaluate the impact of each module of \sysname\ on the model performance. We use \sysname\ to train the model for 150 communication rounds, and record the AP in Table~\ref{tab:ablation}. Take the AP when IoU is above 0.5 as an example, \sysname\ achieves an AP of 0.731, while \sysname\ without MCE loss, modality imputation with autoencoder, and client selection obtain the AP of 0.707, 0.692, and 0.542, respectively. One may think that the MCE loss and modality imputation only improves the result by small margins, while the client selection is much more effective in significantly improving performance. However, it is worth noting that both MCE loss and modality imputation are indispensable parts: although the lack of the two can be compensated by client selection (which excludes erroneous gradients) to a certain extent, there still are many heterogeneous scenarios that cannot be addressed by client selection alone, such as \newrev{those demonstrated in Figures}~\ref{fig:damage_miss_anno} and~\ref{fig:damage_miss_mod}. The integration of MCE loss and modality imputation, together with client selection, can act as ``belt and braces'' to guarantee the robustness of \sysname\ in \newrev{diversified} heterogeneous scenarios.

\begin{table}[h]
    \vspace{-1ex}
	\centering	
	\small
	\caption{Effects of key \sysname\ parts.}
	\label{tab:ablation}
	\vspace{-1.5ex}
    \begin{tabular}{|l|cccc|}
    \hline
            & \multicolumn{4}{c|}{AP}                                                                                   \\ \hline
            & \multicolumn{1}{l|}{IoU=0.5:0.9} & \multicolumn{1}{l|}{IoU=0.5} & \multicolumn{1}{l|}{IoU=0.65} & IoU=0.8 \\ \hline
    \sysname\ & \multicolumn{1}{c|}{0.461}            & \multicolumn{1}{c|}{0.731}        & \multicolumn{1}{c|}{0.698}         &    0.371    \\ \hline
    w/o MCE & \multicolumn{1}{c|}{0.405}            & \multicolumn{1}{c|}{0.707}        & \multicolumn{1}{c|}{0.660}         &0.212   \\ \hline
    w/o AE  & \multicolumn{1}{c|}{0.396}            & \multicolumn{1}{c|}{0.692}        & \multicolumn{1}{c|}{0.657}         &  0.189   \\ \hline
    w/o CS  & \multicolumn{1}{c|}{0.342}            & \multicolumn{1}{c|}{0.542}        & \multicolumn{1}{c|}{0.523}         &   0.272    \\ \hline
    \end{tabular}
    \vspace{-2ex}
\end{table}

\subsection{Hyper-parameter Evaluation}
\subsubsection{Loss Threshold} \label{sssec:loss_thresh}
\begin{figure}[h]
    \setlength\abovecaptionskip{6pt}
    \vspace{-1.5ex}
	   \captionsetup[subfigure]{justification=centering}
		\centering
		\subfloat[Average precision.]{
		  \begin{minipage}[b]{0.47\linewidth}
		        \centering
			    \includegraphics[width = 0.96\textwidth]{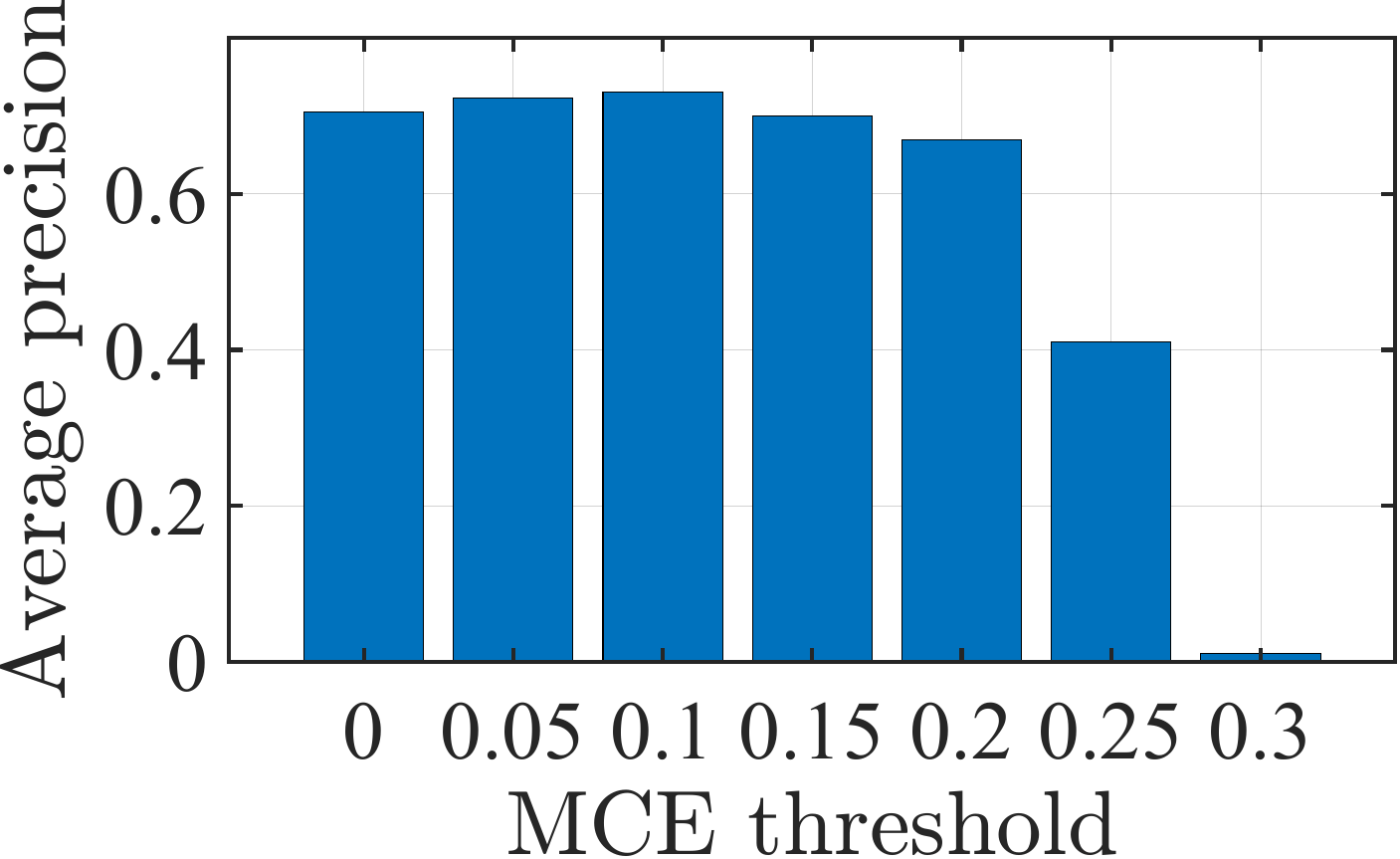}
			\end{minipage}
		    \label{subfig:mce_thresh_ap}
		}
		\subfloat[\brev{Average recall.}]{
		    \begin{minipage}[b]{0.47\linewidth}
		        \centering
			    \includegraphics[width = 0.96\textwidth]{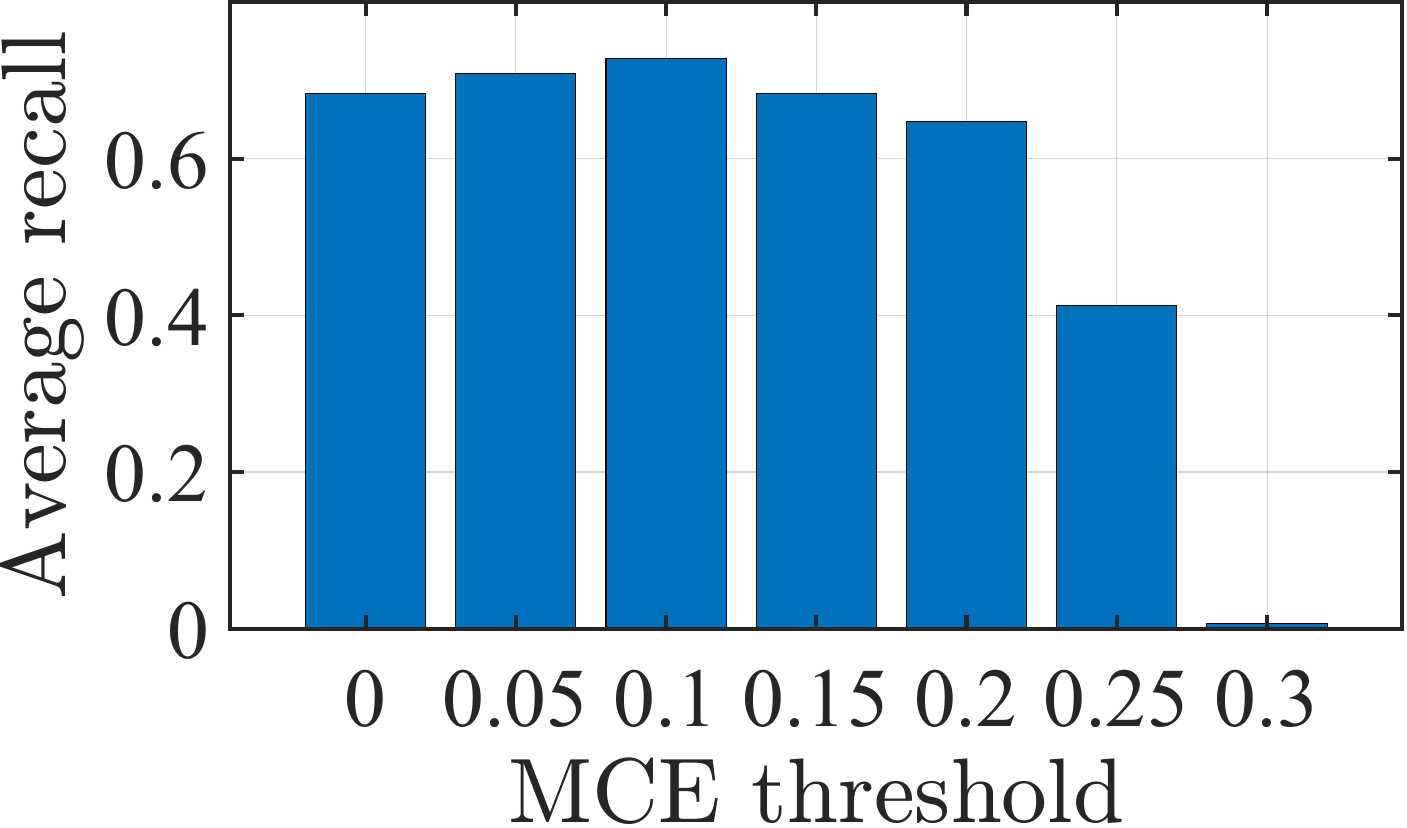}
			\end{minipage}
			\label{subfig:mce_thresh_ar}
		}
		\caption{Impact of the MCE threshold.}
		\label{fig:mce_thresh}
		\vspace{-1.5ex}
\end{figure}

As stated in \S~\ref{sssec:mce}, $p_{\mathrm{th}}$ is a threshold above which we believe that the classifier is more trustworthy than the manual annotations. On one hand, when  $p_{\mathrm{th}}$ is too small, the MCE loss and traditional CE loss are equivalent, and we cannot exclude incorrect gradients induced by missing annotation boxes. On the other hand, many real backgrounds can be mistakenly excluded if $p_{\mathrm{th}}$ is set too large. Therefore, we evaluate the impact of $p_{\mathrm{th}}$ on the \sysname\ performance. As Figure~\ref{subfig:mce_thresh_ap} shows, the AP of vehicle detection increases from 0.7 to 0.73 as $p_{\mathrm{th}}$ increases to 0.1. However, the AP rapidly decreases to around 0 at $p_{\mathrm{th}}=0.3$. \brev{Likewise, a similar trend can be observed in Figure~\ref{subfig:mce_thresh_ar} for AR of the vehicle detection. Overall,} Figure~\ref{fig:mce_thresh} offers a guidance for choosing $p_{\mathrm{th}}$.

\subsubsection{The Number of Selected Clients} 
\begin{figure}[h]
    \setlength\abovecaptionskip{6pt}
    \vspace{-1.5ex}
	   \captionsetup[subfigure]{justification=centering}
		\centering
		\subfloat[Average precision.]{
		  \begin{minipage}[b]{0.47\linewidth}
		        \centering
			    \includegraphics[width = 0.96\textwidth]{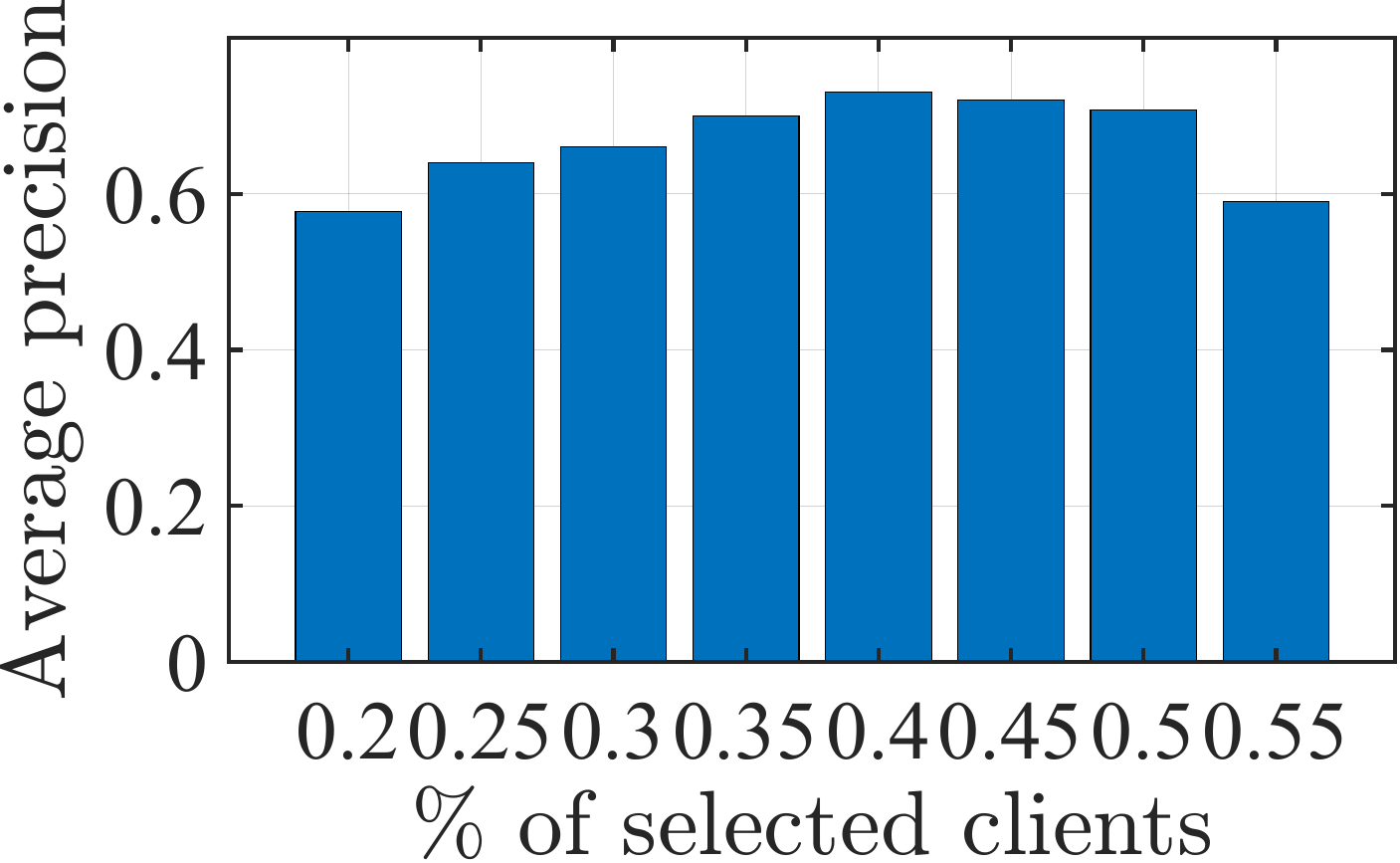}
			\end{minipage}
		    \label{subfig:sel_clients_ap}
		}
		\subfloat[\brev{Average recall.}]{
		    \begin{minipage}[b]{0.47\linewidth}
		        \centering
			    \includegraphics[width = 0.96\textwidth]{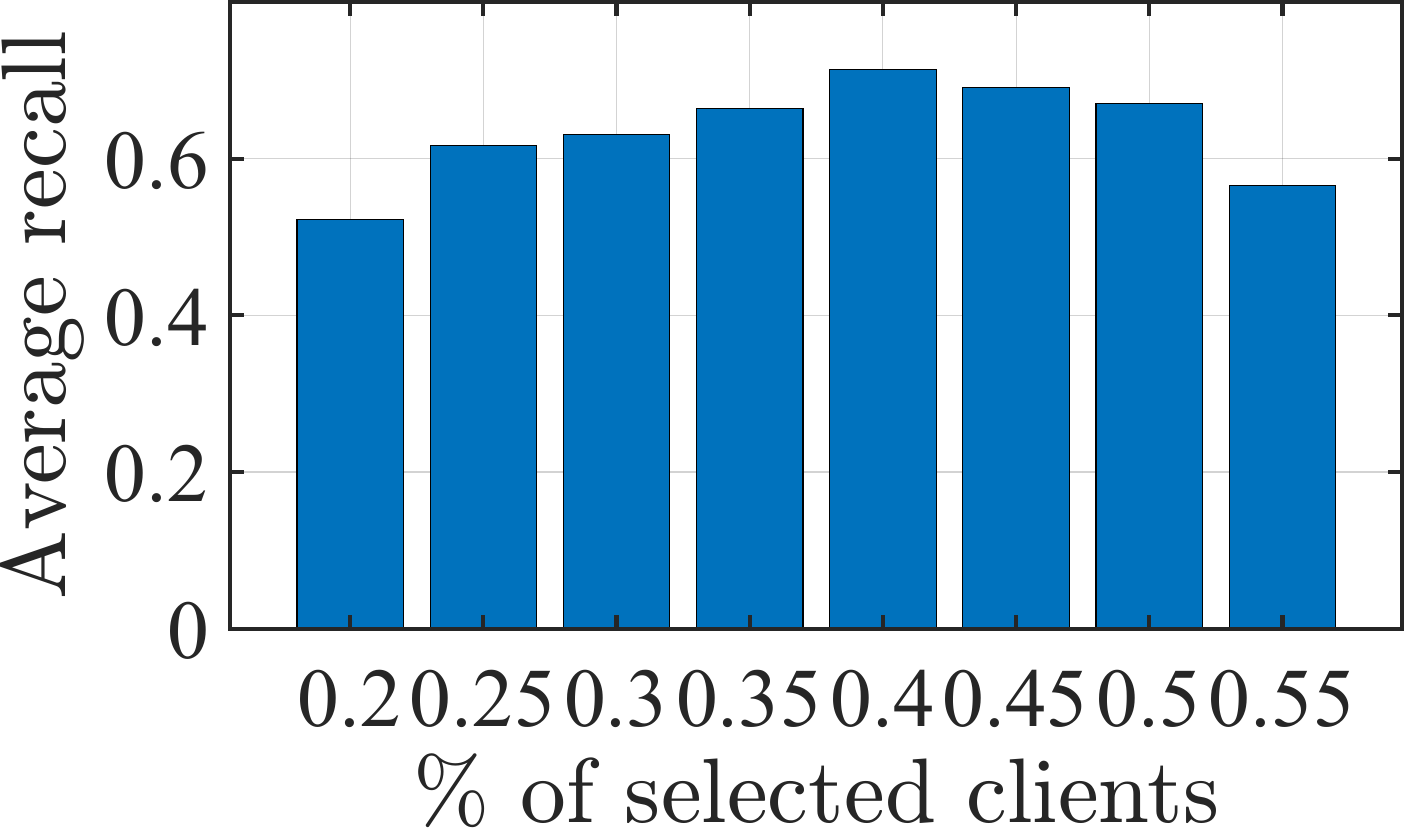}
			\end{minipage}
			\label{subfig:sel_clients_ar}
		}
		\caption{Impact of selected clients percentage.}
		\label{fig:sel_clients}
		\vspace{-1.5ex}
\end{figure}

Another hyperparameter that significantly impacts the performance of \sysname\ is the number of clients selected for model aggregation. On one hand, a small percentage of selected clients could not fully utilize the diverse data collected by different clients and introduce bias into the federated model. On the other hand, if a very large proportion of the clients are selected, we cannot effectively mitigate the detrimental effect caused by diverged local models. Therefore, the number of selected clients balances the tradeoff between utilizing data and excluding diverged models. As Figure~\ref{subfig:sel_clients_ap} shows, the AP of \sysname\ first increases with a greater percentage of selected clients, but starts to drop after the percentage reaches 0.4. The reason is that as the excessive clients are selected for aggregation, the divergence among them will degrade the performance of the federated model. \brev{Furthermore, in Figure~\ref{subfig:sel_clients_ar}, it can be seen that AR of \sysname\ follows a similar trend as AP, and reaches its peak when the percentage of selected clients is 0.4.}%

%% file: 6_related_work.tex
\section{Related Work and Discussion}\label{sec:related_work}

Recent years have witnessed rapid developments in DNN-based OD methods~\cite{redmon2016you, liu2016ssd, lin2017focal, girshick2014rich, girshick2015fast, ren2015faster}. These approaches have been applied to AD~\cite{chen2017multi, yang2018pixor, li20173d, li2016vehicle}. Since most AVs are equipped with multiple sensors (e.g., lidar, radar, and camera), %
\newrev{they become technology} foundations for the OD systems to fully exploit the multimodal data by sensor fusion. Among various sensor fusion schemes, the combination of lidar and another sensor (e.g., radar or camera)~\cite{ku2018joint, xu2018pointfusion, qi2018frustum, chen2017multi, liang2018deep, qian2021robust} is a widely-adopted option due to the complements between each other~\cite{geiger2012we, oxford_robotcar}. One challenge in fusing lidar with other sensors is the unique data structure of lidar, i.e., 3-D point cloud, which is a point set and not compatible with the 2-D matrix in conventional computer vision tasks. One way to overcome this challenge is to employ specially designed DNNs, such as PointNet~\cite{qi2017pointnet}, to directly extract features from point clouds and fuse with other sensing data in the feature space~\cite{xu2018pointfusion}. Another approach is voxelization via transforming the point cloud to 3-D data formats like images, with the height dimension being deemed as image channels. Therefore, the transformed point clouds can be handled by conventional OD-DNNs and fused with other modalities as demonstrated in~\cite{yang2018pixor, simon2019complexer, zhou2018voxelnet, luo2018fast}.

FL~\cite{konevcny2016federated} is a distributed machine learning paradigm that \newrev{transfers only model weights 
instead of explicitly sharing} raw data with the central server. \sysname\ employs FL to enable data crowdsensing without breaching privacy and incurring unaffordable communication cost on AVs. Despite recent FL applications in classification and regression tasks~\cite{konevcny2016federated, li2020federated, li2021fedmask, tu2021feddl,rehman2022federated,so2022fedspace}, applying FL to more sophisticated computer vision tasks such as OD (especially vehicle detection) is far from being exploited. \newrev{In~\cite{jallepalli2021federated}, the authors investigate} the possibility of applying FL to AD applications, and conduct preliminary experiments to verify  privacy protection and convergence speed. FedVision~\cite{liu2020fedvision} proposes an online visual OD platform powered by FL, 
\newrev{but it} focuses more on building and deploying a cloud-based platform, \newrev{without concerning much on FL-related designs.
Fjord \cite{horvath2021fjord} claims to target the data heterogeneity in FL, yet it seems to have missed certain complicated aspects,
such as annotation and modality heterogeneity tackled in \sysname.}

\newrev{While different from existing OD proposals by pioneering federated OD on AVs, \sysname\ is also the first to take into account the effects of all kinds of multimodal heterogeneity for FL-OD on AVs. 
However, \sysname\ still bears one limitation: it stresses on the FL aspect of crowdsensing, pessimistically assuming a finite number of clients unable to provide complete annotations. In other words, we have not considered positive aspects innate to crowdsensing~\cite{ParkGauge-MDM16, WOLoc-INFOCOM17}, such the impact of client incentive~\cite{Truthful-TC16, PostedP-MobiHoc16}. In a future study,} we will extend the design goals of \sysname\ to include designing proper incentives, in order to expand its user base and attract more AV owners to perform collective learning on distributed AV data and \newrev{thus} guarantee \sysname\ service quality.

%% file: 7_conclusion.tex
\section{Conclusion}\label{sec: conclusion}
Taking an important step towards full driving automation, we have proposed \sysname\ in this paper for federated multimodal vehicle detection. Employing a novel loss function, data imputation technique, and client selection strategy, the \sysname\ framework gracefully handles the multimodal data crowdsensed by multiple AV clients, and mines information in the highly heterogeneous data to its maximum, thus releasing its full potential in the vehicle detection task. With extensive experiments under highly heterogeneous scenarios and comparisons with other baselines, we have demonstrated the promising performance of \sysname\ in vehicle detection for autonomous driving. We plan to extend \sysname\ framework to encompass more sensing modalities,
in order to promote its real-life usage and wider acceptance.

Currently, \sysname\ targets on FL-driven vehicle detection, but we are planning to apply FL to other out-vehicle sensing tasks, such as pedestrian detection, lane tracking, and environment semantic segmentation. Moreover, modern vehicles are also equipped with in-vehicle sensing modalities to improve user experience, and we believe FL can help improve the performance of deep analytics upon these modalities too.
Therefore, we are actively exploring the potential of using FL for full vehicle intelligence, particularly for in-vehicle user monitoring (e.g.,~\cite{RF-Net-SenSys20, zheng2020v2ifi, chen2021movifi}); 
this should put us on the right track towards a future with full intelligent transportation.

\section*{Acknowledgement}
We are grateful to anonymous reviewers for their constructive suggestions. This research was support in part by National Research Foundation (NRF) Future Communications Research \& Development Programme (FCP) grant FCP-NTU-RG-2022-015 and MOE Tier 1 Grant RG16/22. We further thank ERI@N and NTU-IGP for supporting the PhD scholarship of Tianyue Zheng.